\documentclass{article}

\usepackage[final]{corl_2026} 

\usepackage[utf8]{inputenc} 
\usepackage[T1]{fontenc}    
\usepackage{hyperref}       
\usepackage{url}            
\usepackage{booktabs}       
\usepackage{amsfonts}       
\usepackage{nicefrac}       
\usepackage{microtype}      
\usepackage{xcolor}         

\usepackage{adjustbox}
\usepackage{tabularx} 
\usepackage{booktabs}
\usepackage{multirow}
\usepackage{hhline}
\usepackage{wrapfig}
\usepackage{subcaption}

\usepackage{algorithm}
\usepackage{algorithmic}
\usepackage{amsmath}
\usepackage{amsfonts}
\usepackage{mathtools}
\usepackage{bm}
\usepackage{amssymb}

\usepackage{cleveref}
\usepackage[table]{xcolor}
\usepackage{makecell}

\usepackage{listings}
\usepackage{xcolor}
\usepackage{inconsolata}
\usepackage{tcolorbox}

\definecolor{mintedbg}{rgb}{0.95,0.95,0.95}
\definecolor{mintedkeyword}{rgb}{0.0, 0.45, 0.65}
\definecolor{mintedstring}{rgb}{0.65, 0.15, 0.15}
\definecolor{mintedcomment}{rgb}{0.25, 0.50, 0.35}
\definecolor{mintednumber}{rgb}{0.5, 0.5, 0.5}

\hypersetup{
    colorlinks,
    linkcolor={blue!50!black},
    citecolor={blue!50!black},
    urlcolor={blue!50!black}
}
\renewcommand{\vec}[1]{\mathbf{#1}}

\title{Real-Time Execution with Autoregressive Policies}

\author{
   Sangkyu Lee$^{1}$ \quad
   Seohyeon Park$^{2}$ \quad
   Tackgeun You$^{1}$ \quad
   Avi Caciularu$^{3}$ \\
   \textbf{Idan Szpektor}$^{3}$ \quad
   \textbf{Hwasup Lim}$^{1}$ \quad
   \textbf{Youngjae Yu}$^{2}$ \\
   $^{1}$Korea Institute of Science and Technology \quad
   $^{2}$Seoul National University \quad 
   $^{3}$Google Research 
   \vspace{2pt} \\
   {\small \url{https://oddqueue.github.io/realfast}}
}

\begin{document}

\maketitle
\begin{abstract}
Real-time execution, enabled by asynchronous inference that ensures both smooth action trajectories and fast reactivity, is critical for realistic deployments of large-scale Vision-Language-Action models.
However, recent work on real-time execution primarily focuses on variants of diffusion policies, even though it is more critical for autoregressive policies given their slower rollout speed in synchronous inference.
In contrast, we demonstrate that autoregressive policies can achieve real-time execution by adjusting the tokenization horizon and applying constrained decoding, thereby guaranteeing strict latency bounds that enable multi-trajectory decoding to maximize performance.
Across simulated and real-world environments, we find that the autoregressive policy consistently outperforms its equivalent-level flow-matching policy counterpart while achieving significantly improved task completion speeds from synchronous inference.
Coupled with the inherent advantages of autoregressive policies, such as faster convergence and better generalizability in instruction-following, these results confirm that autoregressive policies can remain a competitive policy type supporting real-time execution.

\end{abstract}
\section{Introduction}

The era of foundation models, heralded by large language models, has extended to large multimodal models such as Vision-Language models (VLMs), whose capabilities are now being transferred to robotics through Vision-Language-Action models (VLAs).
Although foundation models trained on internet-scale multimodal datasets show great promise for generalizable embodied reasoning~\citep{driess2023palm}, bridging this capability to robot manipulation~\citep{zitkovich2023rt, kim2024openvla, black2024pi_0, black2025pi_, pertsch2025fast, kim2025fine, gr00tn1_2025, zheng2025x, cheang2025gr, internvla_a1, zhao2025cotvla, li2024cogact} poses a significant challenge due to the inherent properties of large-scale parameterized models.
The very engine of their remarkable performance, parameter sizes, introduces inference latency, and flowing time and environments do not wait for their \textit{pausing} of actions during inference in real-world robot deployment.
This sacrifices not only reactivity to environmental changes but also the speed of task completion, since increased robot idle time during inference directly translates into slower rollout speed.

The root cause of this pausing behavior lies in the synchronous inference widely adopted by default for VLAs, or the \textit{think-then-move} approach, where the robot executes actions after policy inference because the policy latency of VLAs generally exceeds the expected command interval of the robot controller in realistic scenarios.
Therefore, the fundamental solution lies in asynchronous inference~\citep{Zhao-RSS-23, liu2025bidirectional, shukor2025smolvla}, or the \textit{think-while-moving}~\citep{xiao2020thinking} approach, ensuring the robot continuously receives actions from the policy; however, this alone does not guarantee satisfactory rollout results.
Without proper constraints, simply maintaining a continuous action stream may cause the policy to ignore the actions of the endpoint robot or fail to react to environmental changes with sufficient frequency.

Investigating asynchronous inference methods that guarantee these constraints, thereby overcoming the inherent latency limitations of VLAs, is the core motivation of recent work aiming to achieve \textit{real-time execution} with VLA policies~\citep{black2026real, sendai2025leave, tang2025vlash, black2025training}.
However, this series of works mainly focuses on VLAs based on the variants of diffusion policy~\citep{chi2025diffusion, hoeg2024streaming} and does not consider the dedicated method for \textit{autoregressive policies}~\citep{zitkovich2023rt, kim2024openvla, pertsch2025fast} that experience slower rollout speed under synchronous inference due to relatively high latency inherent in sequential decoding.
Nevertheless, investigating real-time execution with autoregressive policies is worthwhile, as it can offset the major weakness of slow rollout speeds.
This can position autoregressive policies as a competitive policy type in VLAs, which offer fast convergence and better generalizability in instruction-following~\citep{pertsch2025fast} at the expense of a relatively slower inference frequency while maintaining fast rollout speed on deployment.

\begin{figure}[t]
    \centering
    
    \includegraphics[width=\textwidth]{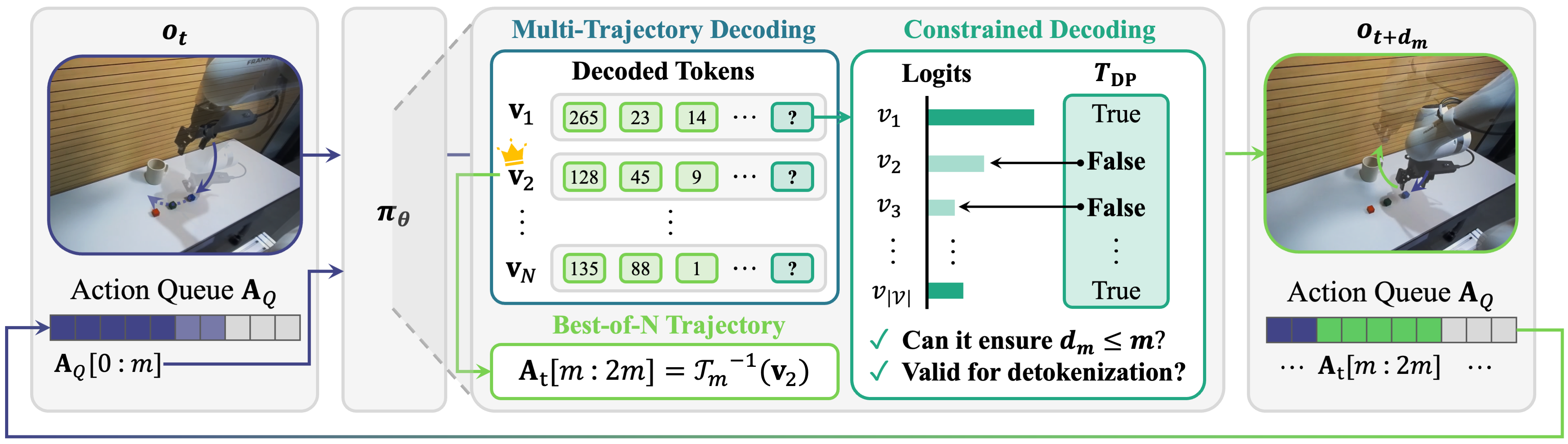}
    
    \vspace{-2pt}
    \caption{\textbf{Overview of Real-Time Execution with Autoregressive Policies.} Given observation $\vec{o}_t$ and action prefix $\vec{A}_Q[0:m]$, $\pi_\theta$ asynchronously repeats multi-trajectory decoding and constrained decoding to guarantee detokenization and latency constraint $d_m \le m$ for avoiding pausing behavior.}
    \label{fig:overview}

\end{figure}

In this work, we present our approach to achieving successful real-time execution with autoregressive policies, as outlined in \Cref{fig:overview}.
Specifically, we demonstrate how to tailor the existing autoregressive policy for supporting real-time execution in four steps: 1) selecting a sufficient action horizon $H = 2m$ to minimize the number of tokens to be decoded for maintaining reasonable latency, 2) applying tokenization on each $m$ horizon rather than the full action horizon $H$ to enable conditioning on action chunks to be executed in the action queue~\citep{shukor2025smolvla}, 3) adopting constrained decoding to guarantee decoding to be completed within the delay $d_m \le m$, and 4) employing multi-trajectory decoding to maximize the performance by fully utilizing idle computation time arising from synchronization.

Throughout the experiments conducted in the simulated environment using LIBERO~\citep{liu2023libero} and the real-world environment using DROID~\citep{khazatsky2024droid}, we confirm that $\mathbf{\pi_0}\textbf{-REALFAST}$, simply fine-tuned from $\pi_0\text{-FAST}$~\citep{pertsch2025fast} with our approach, sufficiently outperforms $\pi_0$~\citep{black2024pi_0} with real-time action chunking (RTC)~\citep{black2026real} in real-time execution, which is the equivalent-level counterpart VLA policy based on flow-matching policy.
Furthermore, it achieves task success rates and rollout speeds that approach those of the successor VLA policy, $\pi_{0.5}$~\citep{black2025pi_}.
We further find that the original advantage of autoregressive policies observed under synchronous inference persists in real-time execution, underscoring the strong motivation for pretraining autoregressive policies to natively support real-time execution.

\section{Related Work}
\paragraph{Real-time Execution with Vision-Language-Action Models}

The transition of VLMs as a foundation model for robotics, VLAs, has emerged as a dominant paradigm~\citep{zitkovich2023rt, kim2024openvla, black2024pi_0, black2025pi_, pertsch2025fast, kim2025fine, gr00tn1_2025, zheng2025x, cheang2025gr, internvla_a1, zhao2025cotvla, li2024cogact}. 
Modern VLAs can be categorized into two groups according to the decoding strategy: 1) autoregressive policies with sequential decoding~\citep{zitkovich2023rt, kim2024openvla, pertsch2025fast}, and 2) continuous action policies, generally variants of diffusion policy~\citep{chi2025diffusion, hoeg2024streaming}, with parallel decoding~\citep{black2025pi_, kim2025fine, gr00tn1_2025, zheng2025x, cheang2025gr, internvla_a1, shukor2025smolvla}.
In recent VLAs, the action chunking policy~\citep{Zhao-RSS-23} for open-loop control is widely assumed, since the inherent inference latency of large-scale VLMs leads to low control frequency and coarse trajectories in closed-loop control~\citep{zitkovich2023rt}.
Here, a series of works address the real-time execution of VLAs, which is asynchronous inference~\citep{Zhao-RSS-23, liu2025bidirectional, shukor2025smolvla}, ensuring both the smoothness in actions and the reactivity, by inference-time inpainting~\citep{black2026real}, correction heads~\citep{sendai2025leave}, or training-time conditioning on the action prefix~\citep{tang2025vlash, black2025training}.

\paragraph{Action Tokenization in Autoregressive Policies}
Autoregressive policies in VLAs rely on the choice of tokenization method to discretize continuous robot actions for next-token prediction.
Early approaches~\citep{zitkovich2023rt, kim2024openvla} adopt a naive binning strategy~\citep{brohan2022rt} for tokenization, discretizing action values across dimensions into 256 bins. However, this yields excessively long token sequences to be decoded, making autoregressive policies unrealistic for high-frequency control.
In this sense, FAST~\citep{pertsch2025fast} proposes a tokenization scheme operating in the discrete cosine transform (DCT) space to compress temporal redundancy and demonstrates that the autoregressive policy can offer faster convergence and better generalizability in instruction-following.
There are several works influenced by FAST that use supervision in the DCT space for VQ tokenization~\citep{liu2025faster}, consider B-spline rather than DCT basis~\citep{zhou2026beast}, or preserve well-definedness that is compromised during the BPE tokenization~\citep{liu2026oatorderedactiontokenization}.
\section{Real-Time Execution with Autoregressive Policies}

\begin{figure}[t]
    \centering

    \includegraphics[width=\textwidth]{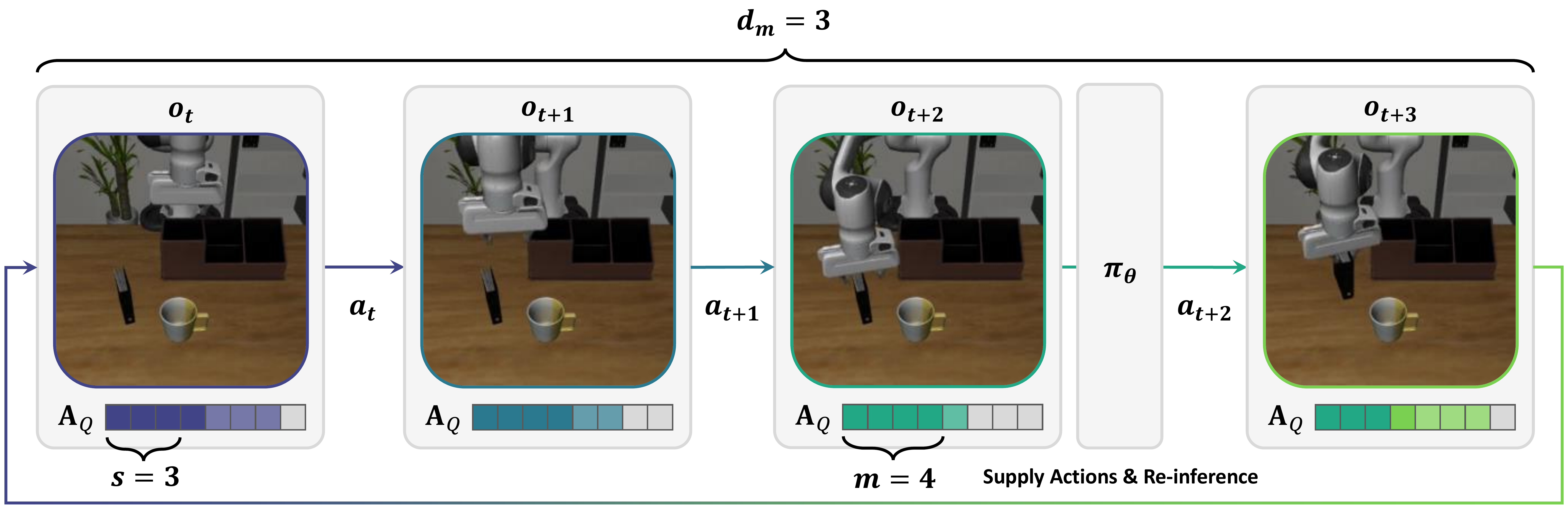}
    
    \vspace{-2pt}
    \caption{\textbf{Example of meeting the necessary condition for real-time execution.} Here, we show the $m=4, s=3$ case where the policy infers with a frequency of $d_m = 3$ with $m=4$ prefix of $\vec{A}_Q$.}
    \label{fig:realtime_requirement}

\end{figure}

\subsection{Decoupling Requirements for Achieving Real-time Execution}

We first clarify the definition of real-time execution in VLAs to understand its requirements, following the formulation in RTC~\citep{black2026real}.
Formally, we assume an action chunking policy $\pi_\theta(\vec{A}_t | \vec{o}_t)$~\citep{Zhao-RSS-23}, where $t$ is a controller timestamp, $\vec{o}_t$ is an observation, $\vec{a}_t$ is an action, $\vec{A}_t \coloneqq [\vec{a}_t, \vec{a}_{t+1}, \cdots, \vec{a}_{t+H-1}]$ is a chunk of future action, and $H$ is an action chunk horizon.
Here, we regard the controller as a \textit{client} that persistently requests $\vec{a}_t$ at each $t$ from the \textit{server} serving $\pi_\theta$, with a sampling period $\Delta t$ defined by $t$.
In general, real-world deployment of policy inference is constrained by latency $\tilde{\delta}_{t, L} \coloneqq \delta_{t, L} + \epsilon_t$, where $\delta_{t, L}$ denotes policy latency for predicting an $L$-length action chunk and $\epsilon_t$ denotes external latency such as network and sensory latency.
Since $\tilde{\delta}_{t, H} >  \Delta t > 0$ holds in realistic deployment~\citep{black2026real}, synchronous inference with open-loop control induces a \textit{pausing} behavior whenever all actions of $\vec{A}_t$ are consumed by the client, which significantly sacrifices the rollout speed of the endpoint robot.

The problem of pausing behavior in synchronous inference arises because the server cannot provide $\vec{a}_t$ to the client at some $t$ for synchronization.
To resolve the pausing behavior, any action request from the client must be fulfilled by $\vec{a}_t$ that is not a predefined \textit{no-op} action halting the robot.
In this sense, we define real-time execution as satisfied whenever the client can consume $\vec{a}_t$ produced by the server for any $t$.
This is analogous to the producer-consumer problem~\citep{dijkstra1968cooperating}, shifting the focus from deterministic response within a strict $\Delta t$ as required in closed-loop control.
In this context, the necessary condition for achieving real-time execution is decoupled from the reactivity or smoothness of the action trajectory.
That is, achieving real-time execution depends only on a deployment strategy that maintains a continuous stream of actions $\vec{a}_t$, making the latency unnoticeable to the client.

One solution for this problem is to introduce an action queue $\vec{A}_Q$ on the client side that chronologically stores chunked actions provided by the server~\citep{shukor2025smolvla}, as depicted in \Cref{fig:realtime_requirement}.
We denote $d_m \coloneqq \lfloor \tilde{\delta}_{t, m} / \Delta t \rfloor $ as the delay that the controller would consume actions under the given execution horizon $s$ during $\tilde{\delta}_{t, m}$, assuming $\vec{o}_t$ is provided along with the consumption of $\vec{a}_{t-1}$~\citep{black2026real}.
If $ d_m \le m \le \lfloor H/2 \rfloor$ is satisfied, we can replace the remaining actions after the consumption of $\vec{A}_Q[0:m]$ with new actions $\vec{A}_t[m:2m]=[ \vec{a}_{t + m}, \vec{a}_{t + m + 1}, \cdots, \vec{a}_{t + 2m - 1} ]$ through asynchronous inference in the frequency of $\delta_{t,m}$.
From the client's perspective, $\vec{A}_Q$, which is the direct source of the action, is not empty for any $t$, so the necessary condition of real-time execution is satisfied.
Therefore, we can achieve real-time execution through asynchronous inference with an appropriate $m$, which defines \textit{non-modifiable} action chunk horizon for prefix of $\vec{A}_Q$ according to $d_m$ with the assumption that $\epsilon_t$ is predictable.

Trivially, the fact that achieving real-time execution depends solely on the choice of $m$ according to $d_m$, regardless of the inference frequency or the smoothness of the action trajectory, implies that real-time execution does not necessarily guarantee successful deployment beyond faster rollout speed.
Here, recent work on real-time execution~\citep{black2026real, sendai2025leave,tang2025vlash, black2025training} focus on guaranteeing these aspects as common motivations with assumption that $m = d_m$: 1) maintain $\delta_{t, m}$ to be sufficiently small, so that the actions from $ \vec{A}_Q $ that the controller consumes are frequently synchronized with the environment changes that $\pi_\theta$ receives, and 2) ensure the trajectory formed by $ \vec{A}_t[m:2m] $ is smoothly connected with the trajectory formed by $\vec{A}_Q[0:m]$, to prevent discontinuity between consecutive action chunks.

\subsection{Revisiting Real-time Execution for Autoregressive Policies}
\label{sec:method_analysis}

Meanwhile, autoregressive policies based on \textit{next-token prediction} are out of scope in recent work for real-time execution~\citep{black2026real, sendai2025leave, tang2025vlash, black2025training}, due to the naturally high $\delta_{t, H}$ induced by sequential decoding compared to parallel decoding of diffusion policies~\citep{chi2025diffusion, hoeg2024streaming}, even with improved tokenization methods such as FAST~\citep{pertsch2025fast}.
Paradoxically, real-time execution, which can accelerate the rollout speed, is more critical for autoregressive policies and thus warrants investigation, since a higher $\delta_{t, H}$ also implies a slower rollout speed in synchronous inference due to longer pausing times. 
Furthermore, the natural advantages of autoregressive policies, such as faster convergence and better generalizability in instruction-following, along with their high architectural compatibility with VLMs~\citep{pertsch2025fast}, make it worthwhile to examine whether reasonable real-time execution can offset the disadvantage.

\begin{figure}[t]
  \vspace{-2pt}
  \centering
  \begin{subfigure}{0.67\columnwidth}
    \centering
    \includegraphics[width=\columnwidth]{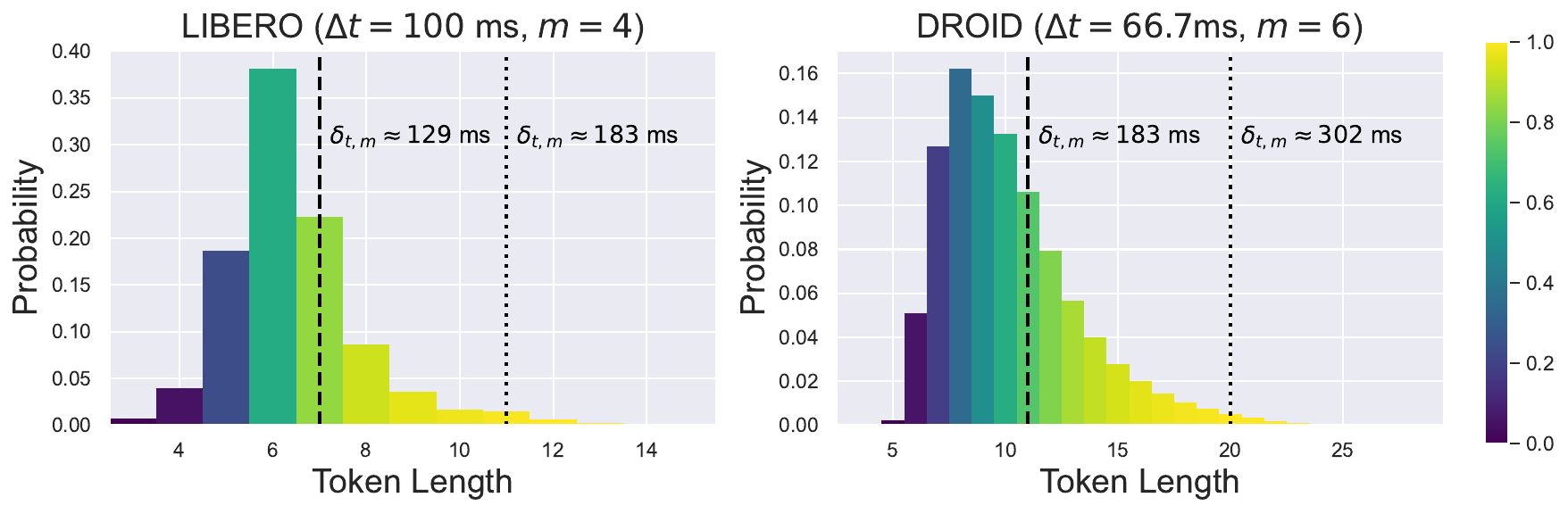}

    \vspace{-4pt}
    \caption{}
    \label{fig:token_length}
  \end{subfigure}
  \hfill
  \begin{subfigure}{0.312\columnwidth}
    \centering
    \includegraphics[width=\columnwidth]{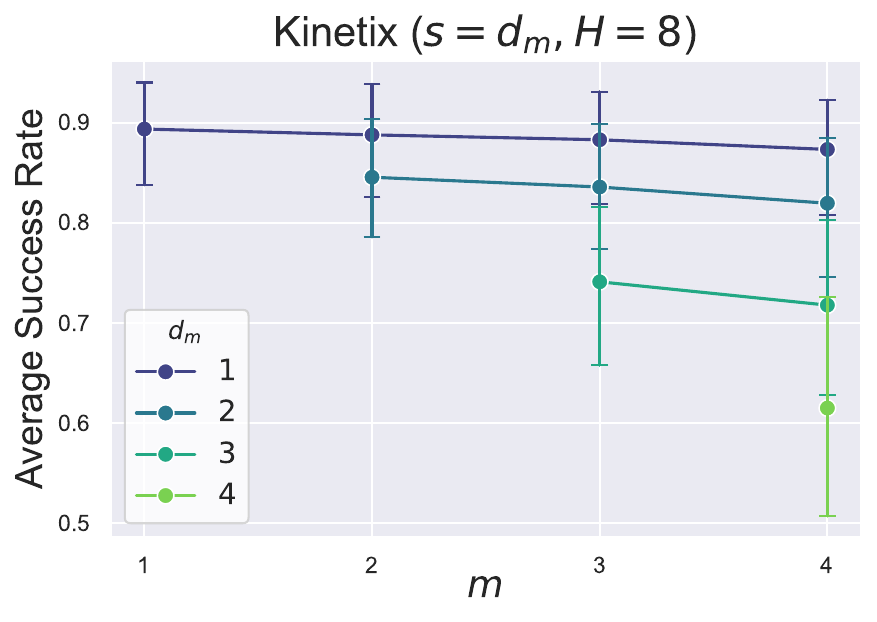}
    
    \vspace{-4pt}
    \caption{}
    \label{fig:kinetix}
  \end{subfigure}
  
  \vspace{-2pt}
  \caption{\textbf{Analysis of factors influencing the real-time execution with autoregressive policies.} (a) Token length distribution of FAST ~\cite{pertsch2025fast} in the LIBERO~\cite{liu2023libero} and DROID~\cite{khazatsky2024droid} datasets. (b) Performance changes of RTC~\cite{black2026real} under $d_m \le m$ in the Kinetix~\cite{matthews2025kinetix} benchmark with 90\% confidence intervals.}
  
\end{figure}

The necessary condition for real-time execution reveals that we have a free choice of $m$, which is directly related to the tokenization horizon $m \cdot \Delta t$ in autoregressive policies.
In other words, we can control $\delta_{t, m}$ by reducing the number of tokens to be decoded, which governs $\delta_{t, m}$, since what we need is to maintain $ d_m \le m \le \lfloor H/2 \rfloor$, not the length of $H$ itself.
In \Cref{fig:token_length}, we present the token length distribution of $400$ ms action chunks from LIBERO~\citep{liu2023libero} and DROID~\citep{khazatsky2024droid} datasets using the FAST+ tokenizer used for $\pi_0\text{-FAST}$~\citep{pertsch2025fast}.
We also measure $\delta_{t, m}$ of $\pi_0\text{-FAST}$ to cover the average token length and $99\%$ of token lengths in each dataset, using a workstation with an RTX 4090 GPU.
We find that the autoregressive policy can achieve a reasonable average $\delta_{t, m}$ for securing reactivity during real-time execution~\citep{black2026real, black2025training, torne2026mem}, even in the absence of a lightweight action expert module as in $\pi_0$~\citep{black2024pi_0}, which shows $\delta_{t, m} \approx 70$ ms in our workstation when using RTC and 5 denoising steps~\citep{black2026real}.

However, this also reveals that $\delta_{t, m}$ cannot be treated as a static value when we consider autoregressive policies with a recent tokenization scheme.
In other words, $m$, the length of the non-modifiable prefix length of $\vec{A}_Q$, and $d_m$, which governs the inference frequency for updating $\vec{A}_Q$ might no longer coincide because $\delta_{t, m}$ varies due to variable lengths of token according to the choice of tokenization scheme so that the inference frequency of the $\pi_{\theta}$ is no longer cyclic during asynchronous inference.
This can lead to a scenario of $d_m \le s \le m$ that is not investigated in recent work~\citep{black2026real, sendai2025leave,tang2025vlash, black2025training}, assuming $d_m \le m \le s$, where the execution horizon $s$ can be regarded as the horizon over which the controller consumes actions without permitting new inference~\citep{black2026real}.
Thus, we need to investigate the impact on performance when the non-modifiable action horizon exceeds the inference frequency.

In \Cref{fig:kinetix}, we track the performance changes of the RTC in the Kinetix benchmark~\citep{black2026real, matthews2025kinetix} according to the change of $d_m$ and $m$, extending the constraint to $d_m = s \le m$.
We confirm that the factor that has a significant impact on performance is $d_m$, which is dominated by the inference frequency $\delta_{t, m}$ in asynchronous inference, while the non-modifiable prefix length $m$ has a relatively small impact under the same policy.
This observation supports the claim that recent autoregressive VLA policies can still be the type of policy for successful real-time execution, since they may be restricted to follow $d_m \le m$ while still achieving a reasonable average $\delta_{t, m}$.
Therefore, we can summarize the real-time execution with autoregressive policies as 1) selecting an appropriate $m$ to secure a reasonable average $\delta_{t, m}$, 2) ensuring continuity with the preceding actions in $\vec{A}_{Q}$, and 3) strictly guaranteeing $d_m \le m$ considering the worst-case of $\delta_{t, m}$ while maximizing the performance.

\subsection{Tailoring Autoregressive Policies for Real-time Execution}
\label{sec:method_tailoring}

To ensure the continuity between non-modifiable action chunks and predicted action chunks, recent work~\citep{black2026real, sendai2025leave,tang2025vlash, black2025training} treats this as additional conditioning on $\vec{A}_Q[0:m]$ along with $\vec{o}_t$.
In autoregressive policies, it is sufficient to \textit{separate the tokenization of the action chunk for conditioning} and provide it as a prefix.
To reduce redundant prediction of actions not required for satisfying the necessary condition of real-time execution, we assume $H = 2m$.
Then, we change the tokenization $\mathcal{T}_H $ performed on the horizon $ H $ to $\mathcal{T}_{m} $ for each $m$-length action chunks in $\vec{A}_{t} $, so that we can separately tokenize the action chunk corresponding to the horizon of the action chunk to be conditioned:
\begin{equation}
\begin{split}
\pi_\theta(\mathcal{T}_H(\vec{A}_t)|\vec{o_t}) \rightarrow \pi_\theta(\mathcal{T}_m(\vec{A}_t[0:m])|\vec{o_t}) \cdot \underbrace{\pi_\theta(\mathcal{T}_m(\vec{A}_t[m:2m])|\vec{o_t}, \mathcal{T}_m(\vec{A}_t[0:m]))}_{\text{conditioned on the preceding action chunk } \vec{A}_t[0:m]}.
\end{split}
\end{equation}

Thus, if there exists $\vec{A}_Q[0:m]$ during deployment, we only decode $\mathcal{T}_m(\vec{A}_t[m:2m])$ which corresponds to $\vec{A}_t[m:2m]$ conditioned on  $\vec{A}_Q[0:m]$, by providing $\vec{A}_Q[0:m]$ to the policy as a prefix along with $\vec{o}_t$.
This modification resolves the domination of $\delta_{t, m}$ by $\delta_{t, H}$ which arises from the tokenizer that decoding the entire $H$-length action chunk is required to obtain $\vec{A}_t[m:2m]$, while enables the conditioning on $\vec{A}_Q[0:m]$ for accomplishing smooth connection between action chunks.

However, decoding $\mathcal{T}_m(\vec{A}_t[m:2m])$ should also follow the restriction $d_m \le m$ to fulfill the necessary condition of real-time execution.
Furthermore, since $\mathcal{T}_m(\vec{A}_t[m:2m])$ is a \textit{token sequence}, we need to ensure that the detokenization $\mathcal{T}_m^{-1}$ yields a valid action chunk~\citep{liu2026oatorderedactiontokenization}.
This is because the $\mathcal{T}_m^{-1}$ is no longer always well-defined in all token sequences, and the failure of $\mathcal{T}_m^{-1}$ needs to supply no-op actions to $\vec{A}_Q$\footnote{For example, it may not satisfy the required coefficient length for the inverse DCT used for $\mathcal{T}^{-1}$ of the FAST. Accordingly, $\pi_0\text{-FAST}$ output actions computed from dataset statistics when $\mathcal{T}^{-1}$ fails in default.}.
Therefore, \textit{constrained decoding}~\citep{zitkovich2023rt} is required for meeting restrictions, as other sequence generation problems with predefined rules~\citep{hokamp2017lexically, post2018fast, deutsch2019general}.
Formally, assume reasonable bound of $\epsilon_t$ exists and define $\delta' \coloneqq \sup \{ \delta \mid \lfloor (\delta + \sup \epsilon_t) / \Delta t \rfloor \le m \}$.
Here, we denote the latency for decoding $L$ length of tokens by $\pi_\theta$ as $\delta_L$ and define $B \coloneqq \max \{ L \in \mathbb{N} \mid \delta_L \le \delta' \}$.
What we want to enforce is that the decoded token sequence $\vec{v} = [v_1, v_2, \cdots, v_l] \in \mathcal{V}^l$ must satisfy $ l \le B$ and $\mathcal{T}_m^{-1}(\vec{v}) \in \mathbb{R}^{m \times d}$ is well-defined, where $d$ is the action dimension and $ \mathcal{V} $ is the action vocabulary.

To maintain generality, we define $C_v$ as the amount that $v$ contributes to the capacity $C$, where $C$ is the requirement for $\mathcal{T}_m^{-1}(\vec{v})$ to be valid detokenization\footnote{$C_{v}$ is the length of the DCT coefficient corresponding to each $v$, and $C = m \times d$ when considering FAST.}.
In other words, a token sequence $\vec{v}$ is detokenizable if and only if it satisfies $C_{\vec{v}} \coloneqq \sum_{i=1}^n C_{v_i} = C$.
In this case, we can interpret decoding a detokenizable token sequence of length at most $B$ as a bounded variant of the coin change problem: forming a value $C$ with at most $B$ coins of coin denomination $C_v$, which can be efficiently resolved by dynamic programming (DP).
Specifically, we construct a boolean DP table $T_{\text{DP}} \in \{0, 1\}^{(B+1) \times (C+1)}$, starting from the base case $T_{\text{DP}}[0, 0] = 1$ and $0$ for otherwise, then recursively updated as follows:
\begin{equation}
\begin{split}
T_{\text{DP}} [i, c] = T_{\text{DP}} [i-1, c] \vee ( \bigvee_{v \in \mathcal{V}} T_{\text{DP}} [i-1, c - C_v] ),
\end{split}
\end{equation}
where $T_{\text{DP}}[i, c'] = 0$ for any $c' < 0$.
Here, each entry $T_{\text{DP}}[i, c]$ indicates whether the remaining capacity $c$ can be fulfilled within at most $i$ decoding steps, and this DP table can be precomputed with efficient vectorized boolean operations before deployment.
At decoding step $j$, we can mask out the logit of new token $v$ which does not satisfy the restriction by checking $T_{\text{DP}}[B-j, C - C_{\vec{v}_{j-1}} - C_v] = 0$ with $O(1)$ memory lookup, where $\vec{v}_{j-1}$ denotes the token sequence decoded in previous steps.

Therefore, we can guarantee $d_m \le m$ with constrained decoding to restrict the upper bound of $\delta_{t, m}$ and repeat the inference asynchronously whenever the decoding ends before $B$ steps with carefully choosing $m$ according to the expected $\epsilon_t$ and $B$ that sufficiently cover the token length distribution. 
Moreover, the fact that it is possible to compute the \textit{exact} likelihood under constrained support with a predictable latency bound implies that we can perform \textit{multi-trajectory decoding} to select the best-of-$N$ trajectory according to the likelihood.
The latency overhead of multi-trajectory decoding remains marginal\footnote{Specifically, we find that decoding $11$ or $20$ tokens covering $99\%$ of token lengths from LIBERO or DROID dataset by $\pi_0\text{-FAST}$ with $N = 2, 4$ increases latency by approximately $4$, $7$ or $8$, $13$ ms, respectively.} when sharing the prefix KV cache~\citep{jang2025verifier} for reasonable $N$, because the overhead comes only from generated KV caches, given the memory-bound nature of decoding in autoregressive models~\citep {pope2023efficiently}.
Therefore, we can adopt multi-trajectory decoding to fully maximize performance by reducing the computation idle time, $ (\tilde{\delta}_{t, m} / \Delta t) - d_m$, that can happen during synchronization.
\section{Experiments}

\subsection{Experimental Setup}

\paragraph{Baselines}
We validate our approach in the \textit{single-arm robot manipulation} setting, treating $\pi_0\text{-FAST}$~\citep{pertsch2025fast} as a representative autoregressive policy. 
We refer to the case where our approach is applied as $\mathbf{\pi_0}\textbf{-REALFAST}$ to distinguish it from the original model.
Throughout the experiment, we fix $m \cdot \Delta t = 400$ ms for $\pi_0\text{-REALFAST}$ to maintain consistency with \Cref{sec:method_analysis}.
We mainly compare our approach with RTC~\citep{black2026real} applied to $\pi_0$~\citep{black2024pi_0} as a fair baseline, which shares the VLM backbone~\citep{beyer2024paligemma} and uses the same pretraining dataset as $\pi_0\text{-FAST}$, but adopting a lightweight flow-matching action expert.
Furthermore, we regard $\pi_{0.5}$~\citep{black2024pi_0} with RTC as a formidable baseline for successful real-time execution, since $\pi_{0.5}$ is trained with additional multimodal web-scale datasets over $\pi_0$ and $\pi_0\text{-FAST}$, and adopts both of the lightweight flow-matching action expert and FAST for faster adaptation~\citep{driess2026knowledge}.

\paragraph{Simulated Environment}
We use \textbf{LIBERO}~\citep{liu2023libero} as the simulated environment, following OpenVLA~\citep{kim2024openvla}.
However, since the LIBERO benchmark does not reflect $\tilde{\delta}_{t, m}$, we introduce $\vec{A}_Q$ that can be fully accessible to the policy but restricted in modifying the $s$-length prefix of $\vec{A}_Q$ to simulate real-time execution while fixing $s$ as the upper bound of $d_m$.
We also assume $\epsilon_t \in [\Delta t, 2 \Delta t, 3 \Delta t] $ to investigate the robustness on external latency, where $\Delta t = 100$ ms.
Here, $\delta_{t, m}$ of $\pi_0\text{-REALFAST}$ is approximately bounded to $183$ ms with constrained decoding under $B=11$ to cover 99\% of the token length of action chunks on the workstation with RTX 4090 GPU.
We sweep $[3k, 6k, 10k, 15k, 30k]$ training steps and only consider $N = 1, 2, 4$ which does not change upper bound $d_m$ during deployment.
We select the best model across different training steps while fixing sampling temperature $\tau=0$, then sweep from $0$ to $1.0$ in increments of $0.1$ and use beam search when $\tau = 0$ for $N > 1$.

\paragraph{Real-world Environment}
We use \textbf{DROID}~\citep{khazatsky2024droid} as the real-world environment for \textit{zero-shot} tabletop manipulation in unseen scenes~\citep{pertsch2025fast} and evaluate policies fine-tuned on the DROID dataset without additional demonstrations.
We configure evaluation suites of 8 tabletop manipulation tasks, with 10 trials per task and a time limit of 40 seconds; a trial is considered a failure if it is not completed within the time limit.
$\pi_0\text{-REALFAST}$ follows $m = 6$ and we set $m = 2$ for $\pi_0$ or $\pi_{0.5}$ with RTC, where approximately $\epsilon_t $ is measured as $9$ ms and $\delta_{t, m} $ follows $70$ or $74$ ms, respectively, with $\Delta t = 66.7$ ms.
Here, $\delta_{t, m}$ of $\pi_0\text{-REALFAST}$ is bounded to approximately $302$ ms with constrained decoding under $B=20$ to cover 99\% of the token length of action chunks.
We use multi-trajectory decoding with $N=4$, and it yields approximately $324$ ms for $\tilde{\delta}_{t, m}$ without increasing the upper bound of $d_m$.
Further details about the implementation and experimental setup are included in the appendix.

\begin{table}[t]
    \centering

    \caption{\textbf{Evaluation results in the LIBERO benchmark when restricted to real-time execution.} We report the average and standard deviation of the task success rate resulting from three different random seeds. We also include baselines without inference delay for reference, denoted as $m=0$.}
    \vspace{2pt}
    \adjustbox{max width=\textwidth}{
        \setlength{\tabcolsep}{4pt}
        \begin{tabular}{llccccccc}
            \toprule
            \textbf{Model} & \textbf{Deployment} & \textbf{LIBERO-Spatial} & \textbf{LIBERO-Object} & \textbf{LIBERO-Goal} & \textbf{LIBERO-Long} & \textbf{Average} \\
            \midrule
            \color{gray} $\pi_0 $~\citep{black2024pi_0} & \color{gray} $m = 0, s = 5$ & \color{gray} 96.8 \% & \color{gray} 98.8 \% & \color{gray} 95.8 \% & \color{gray} 85.2 \% & \color{gray} 94.2 \% \\
            \color{gray} $\pi_{0.5} $~\citep{black2025pi_} & \color{gray} $m = 0, s = 5$ & \color{gray} 98.8 \% & \color{gray} 98.2 \% & \color{gray} 98.0 \% & \color{gray} 92.4 \% & \color{gray} 96.9 \% \\
            \color{gray} $\pi_0\text{-FAST}$~\citep{pertsch2025fast} & \color{gray} $m = 0, s = 5$ & \color{gray} 93.0 \% & \color{gray} 94.0 \% & \color{gray} 91.0 \% & \color{gray} 77.0 \% & \color{gray} 88.8 \% \\
            \midrule
            \multirow{6}{*}{$\pi_0 $ + RTC} & $m = 1, s = 1$ & 97.4 ($\pm$ 0.2) \% & 99.3 ($\pm$ 0.1) \% & 96.7 ($\pm$ 0.2) \% & 82.4 ($\pm$ 0.4) \% & 94.0 ($\pm$ 0.2) \% \\
            & $m = 2, s = 2$ & 96.7 ($\pm$ 1.3) \% & 99.6 ($\pm$ 0.2) \% & 97.7 ($\pm$ 0.8) \% & 87.1 ($\pm$ 0.5) \% & 94.7 ($\pm$ 0.1) \% \\
            & $m = 3, s = 3$ & 95.7 ($\pm$ 0.6) \% & 99.5 ($\pm$ 0.3) \% & 96.9 ($\pm$ 0.1) \% & 86.5 ($\pm$ 0.8) \% & 94.7 ($\pm$ 0.2) \% \\
            \cmidrule(lr){2-7}
            & $m = 4, s = 1$ & 96.8 ($\pm$ 0.4) \% & 99.4 ($\pm$ 0.5) \% & 96.4 ($\pm$ 0.4) \% & 86.7 ($\pm$ 1.1) \% & 95.0 ($\pm$ 0.1) \% \\
            & $m = 4, s = 2$ & 96.1 ($\pm$ 0.8) \% & 99.3 ($\pm$ 0.6) \% & 96.3 ($\pm$ 0.4) \% & 87.4 ($\pm$ 1.1) \% & 94.8 ($\pm$ 0.2) \% \\
            & $m = 4, s = 3$ & 95.7 ($\pm$ 0.6) \% & 99.2 ($\pm$ 0.2) \% & 97.0 ($\pm$ 0.5) \% & 89.2 ($\pm$ 0.9) \% & 95.3 ($\pm$ 0.4) \% \\
            \midrule
            \multirow{6}{*}{$\pi_{0.5} $ + RTC} & $m = 1, s = 1$ & 98.7 ($\pm$ 0.8) \% & 98.4 ($\pm$ 0.3) \% & 97.3 ($\pm$ 0.3) \% & 91.8 ($\pm$ 0.4) \% & 96.5 ($\pm$ 0.1) \% \\
            & $m = 2, s = 2$ & 98.8 ($\pm$ 0.0) \% & 98.7 ($\pm$ 0.2) \% & 98.4 ($\pm$ 0.2) \% & 92.7 ($\pm$ 1.0) \% & 97.2 ($\pm$ 0.2) \% \\
            & $m = 3, s = 3$ & 98.3 ($\pm$ 0.1) \% & 99.0 ($\pm$ 0.2) \% & 97.9 ($\pm$ 0.1) \% & 93.7 ($\pm$ 0.1) \% & 97.2 ($\pm$ 0.1) \% \\
            \cmidrule(lr){2-7}
            & $m = 4, s = 1$ & 98.5 ($\pm$ 0.8) \% & 99.1 ($\pm$ 0.5) \% & 97.3 ($\pm$ 0.3) \% & 93.3 ($\pm$ 0.8) \% & 97.1 ($\pm$ 0.5) \% \\
            & $m = 4, s = 2$ & 98.4 ($\pm$ 0.7) \% & 99.0 ($\pm$ 0.3) \% & 97.4 ($\pm$ 1.2) \% & 93.9 ($\pm$ 0.6) \% & 97.2 ($\pm$ 0.6) \% \\
            & $m = 4, s = 3$ & 98.6 ($\pm$ 0.4) \% & 99.3 ($\pm$ 0.1) \% & 97.1 ($\pm$ 0.5) \% & 93.5 ($\pm$ 0.5) \% & 97.2 ($\pm$ 0.1) \% \\
            \midrule
            \rowcolor{gray!20}
            & $m = 4, s = 2$ & 97.7 ($\pm$ 0.5) \% & 98.9 ($\pm$ 0.5) \% & 95.7 ($\pm$ 0.5) \% & 91.5 ($\pm$ 1.7) \% & 96.0 ($\pm$ 0.7) \% \\
            \rowcolor{gray!20}
            & $m = 4, s = 3$ & 97.7 ($\pm$ 0.8) \% & 98.9 ($\pm$ 0.4) \% & 95.9 ($\pm$ 1.3) \% & 90.0 ($\pm$ 0.2) \% & 95.6 ($\pm$ 0.3) \% \\
            \rowcolor{gray!20}
            \multirow{-3}{*}{$\bm{\pi_0}\textbf{-REALFAST}$} & $m = 4, s = 4$ & 97.2 ($\pm$ 0.4) \% & 98.6 ($\pm$ 0.5) \% & 97.1 ($\pm$ 0.4) \% & 89.9 ($\pm$ 0.3) \% & 95.7 ($\pm$ 0.3) \% \\
            \midrule 
            \multicolumn{7}{c}{\textbf{Test-time Compute Matched}}  \\
            \midrule
            \multirow{2}{*}{$\pi_0 $ + RTC} & $m = 3, s = 3$ & 94.9 ($\pm$ 0.6) \% & 98.7 ($\pm$ 0.5) \% & 96.7 ($\pm$ 1.0) \% & 85.6 ($\pm$ 0.8) \% & 94.0 ($\pm$ 0.3) \% \\
            & $m = 4, s = 3$ & 93.8 ($\pm$ 0.6) \% & 99.1 ($\pm$ 0.3) \% & 95.7 ($\pm$ 0.3) \% & 87.5 ($\pm$ 0.7) \% & 94.0 ($\pm$ 0.2) \% \\
            \midrule
            \multirow{2}{*}{$\pi_{0.5} $ + RTC} & $m = 3, s = 3$ & 98.2 ($\pm$ 0.4) \% & 98.8 ($\pm$ 0.5) \% & 97.6 ($\pm$ 0.9) \% & 93.0 ($\pm$ 0.8) \% & 96.9 ($\pm$ 0.1) \% \\
            & $m = 4, s = 3$ & 98.7 ($\pm$ 0.1) \% & 98.7 ($\pm$ 0.3) \% & 96.9 ($\pm$ 1.0) \% & 93.6 ($\pm$ 1.3) \% & 97.0 ($\pm$ 0.4) \% \\
            \midrule
            \rowcolor{gray!20}
            $\bm{\pi_0}\textbf{-REALFAST}$ & $m = 4, s = 2$ & 97.7 ($\pm$ 0.5) \% & 98.9 ($\pm$ 0.5) \% & 95.7 ($\pm$ 0.5) \% & 91.5 ($\pm$ 1.7) \% & 96.0 ($\pm$ 0.7) \% \\
            \bottomrule
        \end{tabular}
    }

    \vspace{-10pt}
    \label{tab:delay}
\end{table}
\begin{figure}[t]
  \centering
      \begin{subfigure}{0.48\columnwidth}
        \includegraphics[width=\columnwidth]{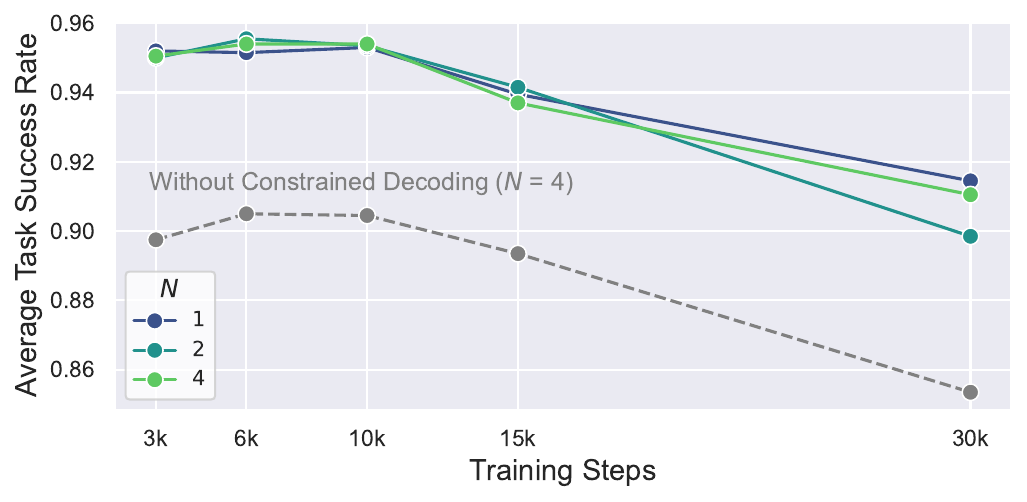}
        
        \vspace{-4pt}
        \caption{}
        \label{fig:convergence}
      \end{subfigure}
      \hfill
      \begin{subfigure}{0.48\columnwidth}
        \includegraphics[width=\columnwidth]{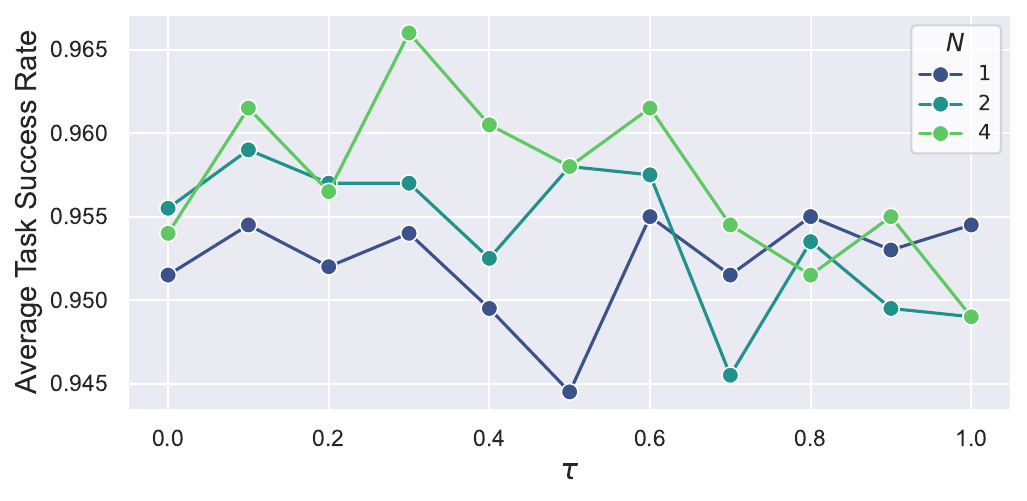}
        
        \vspace{-4pt}
        \caption{}
        \label{fig:temperature}
      \end{subfigure}
      
      \vspace{-2pt}
      \caption{\textbf{Ablation study in the LIBERO environment.} (a) Average task success rate according to the training step. (b) Average task success rate according to $\tau$ of the model with $6k$ training steps.}
      
    \label{fig:droid_results}
\end{figure}

\begin{figure}[t]
  \centering
      \begin{subfigure}{0.48\columnwidth}
        \includegraphics[width=\columnwidth]{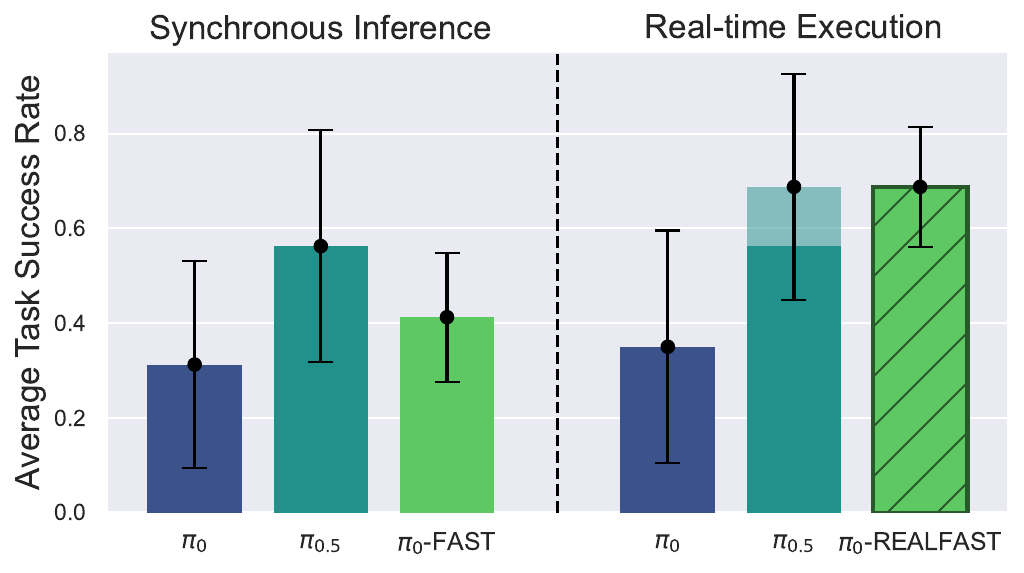}
      \end{subfigure}
      \hfill
      \begin{subfigure}{0.48\columnwidth}
        \includegraphics[width=\columnwidth]{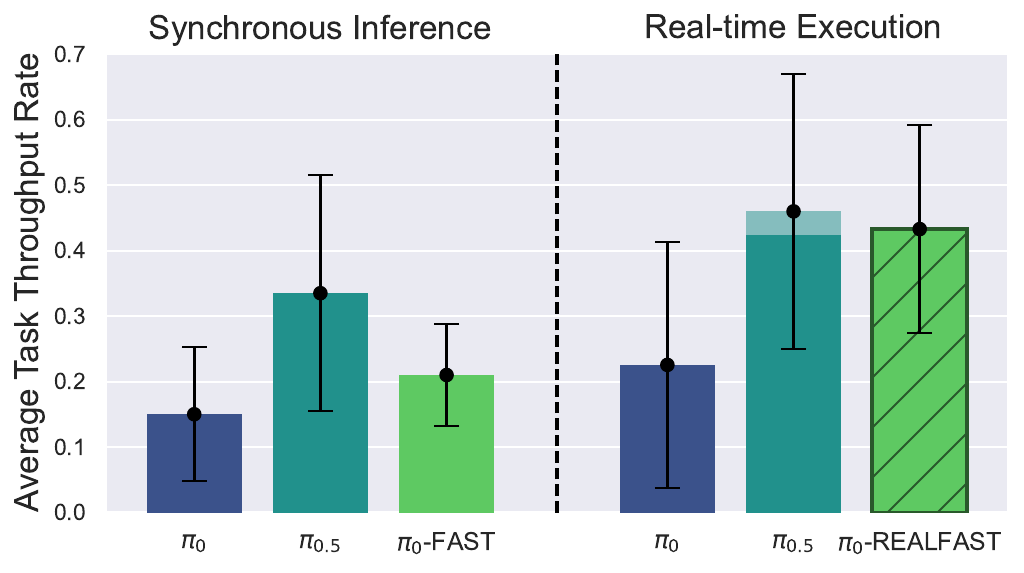}
      \end{subfigure}
      
      \caption{\textbf{Evaluation results on zero-shot DROID deployment.} We report 90\% confidence intervals.}
      \label{fig:droid_quantitative}
      
      \vspace{-8pt}

\end{figure}
\begin{figure}[t]
    \centering
  
    \includegraphics[width=\textwidth]{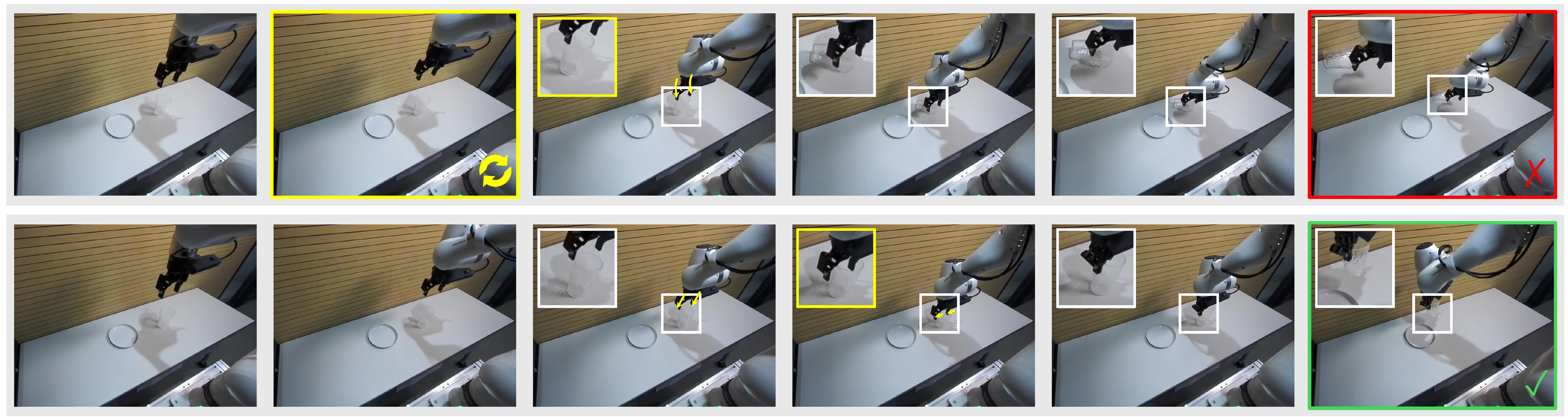}

    \vspace{-2pt}

    \caption{\textbf{Qualitative comparison between $\bm{\pi_0}\textbf{-FAST}$ (top) and $\bm{\pi_0}\textbf{-REALFAST}$ (bottom).} We highlight events when object contact occurs in yellow, given instruction \textit{"place the cup on the plate"}.}
    \label{fig:droid_qualitative}

\end{figure}

\subsection{Experimental Results}
\label{sec:experimental_results}

\paragraph{Effect of Non-modifiable Action Horizon}

In \Cref{tab:delay}, we compare the task success rate in the LIBERO benchmark under real-time execution with varying $\epsilon_t$.
We also evaluate RTC with $d_m \le m = 4$ for tabletop manipulation beyond the 2D Kinetix environment to understand the effect of $d_m \le m$, as discussed in \Cref{sec:method_analysis}.
We observe that $\pi_0\text{-REALFAST}$ consistently surpasses $\pi_0$, yet slightly inferior compared to $\pi_{0.5}$.
Notably, policies deployed with $d_m \le m$ consistently maintain task success rate in changes of $s$, whereas the RTC with $d_m = m$ exhibits degeneration at \textit{lower inference latency} in both $\pi_0$ and $\pi_{0.5}$.
Although RTC reports a similar phenomenon for injected latency~\citep{black2026real}, we infer that increasing $m$ likely prevents degeneration, since we find that $d_m \le m$ remains robust at the same inference frequency.
We further check whether the small test-time compute of $\pi_0$ and $\pi_{0.5}$ yields inferior results, given that multi-trajectory decoding can be regarded as test-time compute scaling.
We match the decoding compute of $\pi_0$ and $\pi_{0.5}$ by increasing denoising steps to 30, requiring about 39 GFLOPs as in $\pi_0\text{-REALFAST}$ with $N = 4$, but still observed similar performance superiority.
Therefore, we can infer that aggressively modifying near-future actions in $\vec{A}_Q$ without appropriate test-time compute scaling does not necessarily enable successful real-time execution.

\paragraph{Advantage of Constrained Decoding and Multi-trajectory Decoding}
\Cref{fig:convergence} tracks the change in the average task success rate at $\tau = 0$ with constrained decoding applied according to the training step, and the case limited in decoding length $B = 11$ but without constrained decoding.
Here, when constrained decoding is not applied, the best-of-$N$ trajectory is selected by likelihood but is limited to candidates of a detokenizable token sequence.
We find that $\pi_0\text{-REALFAST}$ quickly reaches its best performance within 6k training steps, approximately 6 epochs, consistent with the fast convergence of autoregressive policies~\citep{pertsch2025fast}, and that excessive fine-tuning beyond this point, as well as the removal of constrained decoding, both lead to degradation.
In \Cref{fig:temperature}, we further investigate the temperature sampling strategy compared to greedy decoding and beam search using the model trained with $6k$ training steps.
We confirm that the strategically configured multi-trajectory decoding strategy tailored to the deployment environment serves as an effective method for maximizing performance.

\paragraph{Quantitative Evaluation in Zero-shot Deployment}

In \Cref{fig:droid_quantitative}, we show the average task success and throughput rate of $\pi_0\text{-REALFAST}$, $\pi_0$, and $\pi_{0.5}$ in real-time execution on evaluation suites for zero-shot deployment in the DROID environment, along with the synchronous inference results of $\pi_0$, $\pi_{0.5}$, and $\pi_0\text{-FAST}$ with $ s = 6 $.
Here, we illustrate the results of RTC with $d_m \le m = 6$ using higher opacities, and we measure task throughput rate as the ratio of the remaining time until success to the maximum time limit, with 0 for failure to reflect the rollout speed.
We observe that $\pi_0\text{-REALFAST}$ significantly surpasses the base model with synchronous inference, $\pi_0\text{-FAST}$, while outperforming $\pi_0$ and showing performance comparable to $\pi_{0.5}$.
We can argue that the better generalizability in instruction-following demonstrated by autoregressive policies~\citep{pertsch2025fast} is preserved in real-time execution, considering the superior performance of $\pi_0\text{-FAST}$ in synchronous inference compared to $\pi_0$.
Meanwhile, we confirm that $\pi_{0.5}$ operates regardless of choice $d_m \le m$ and $d_m = m$, but $\pi_{0}$ completely fails in zero-shot deployment under $d_m = m$, implying that policy robustness should be prioritized over reducing latency for faster update of near-future actions in $\vec{A}_Q$.

\paragraph{Qualitative Analysis of Real-time Execution}

In \Cref{fig:droid_qualitative}, we perform a qualitative comparison between $\pi_0\text{-FAST}$ and $\pi_0\text{-REALFAST}$, since the improvement in average task success rate is more pronounced in $\pi_0\text{-REALFAST}$ from synchronous inference compared to $\pi_{0}$ and $\pi_{0.5}$.
We observe that in synchronous inference, collisions with the target object can easily occur when inference occurs at a distance from it.
In particular, because of the inherent nature of synchronous open-loop control, the policy cannot adjust actions during execution, resulting in \textit{overshoot} in approaching the object.
In contrast, we observe that $\pi_0\text{-REALFAST}$ avoids collisions through fine trajectory adjustments during the approach, even when the initial approach direction is suboptimal.
Specifically, the policy completes the grasp by quickly modifying its trajectory just before contact, while maintaining the initial path would likely cause a collision with the rim when grasping a cup.
This demonstrates that real-time execution can significantly improve performance over synchronous inference, even in relatively static environments, by enabling frequent adjustments that allow precise object manipulation.
\section{Conclusion}

In this work, we have demonstrated real-time execution with autoregressive policies by supporting asynchronous inference conditioned on preceding action chunks while guaranteeing the prediction of diverse action trajectories before inducing pausing behavior at the endpoint robot.
Experimental results across simulated and real-world environments reveal that the current stage of autoregressive VLA policty is already affordable for real-time single-arm robot manipulation, with sufficient consideration of inference latency.
In particular, we find that the autoregressive policy can outperform its equivalent-level counterpart with lower inference latency in real-time execution, and reducing policy latency does not necessarily lead to better deployment.
We hope that this finding provides an opportunity to shed light on the trade-off between policy robustness and inference latency, which has been overlooked in recent work on VLAs that focused on reducing latency to secure high reactivity.
\section{Limitations and Discussions}

\begin{figure*}[t]
\centering

\begin{subfigure}[b]{0.49\textwidth}
    \centering
    \includegraphics[
        width=\linewidth
    ]{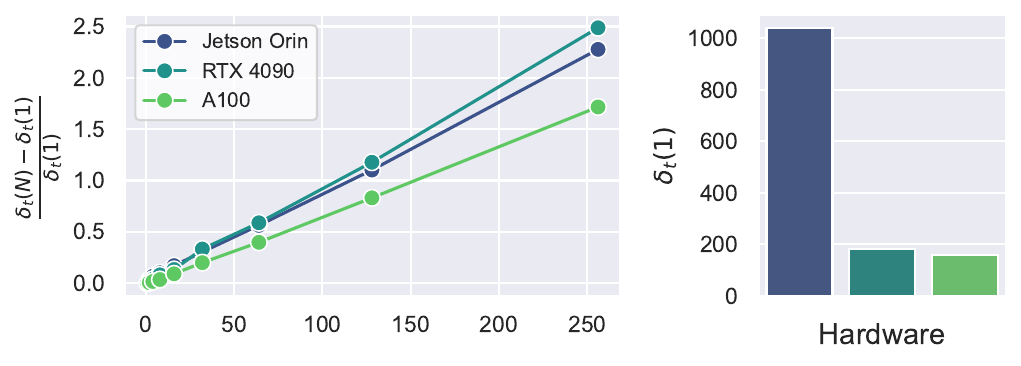}
    \caption{}
    \label{fig:droid_hardware}
\end{subfigure}
\hfill
\begin{subfigure}[b]{0.49\textwidth}
    \centering
    \resizebox{\linewidth}{!}{%
        \begin{tabular}{cclc}
            \toprule
            \textbf{Task Scene}
            & \textbf{Inference}
            & \textbf{Model}
            & \textbf{Success Rate} \\
            \midrule

            \multirow{6}{*}{%
                \includegraphics[width=2.6cm]
                {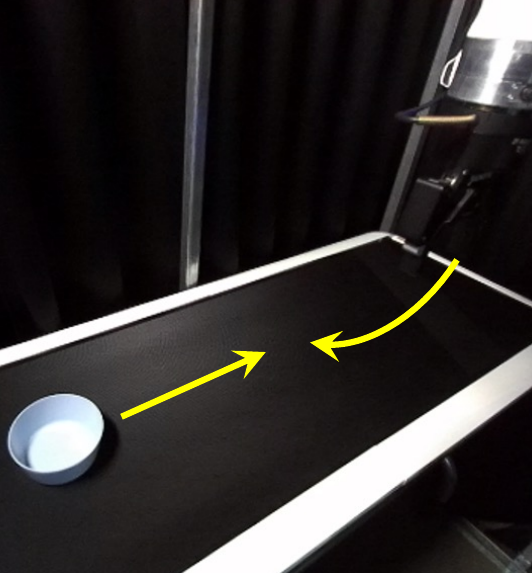}
            }
            & \multirow{3}{*}{Synchronous}
            & $\pi_0$
            & 55\% (11/20) \\

            & & $\pi_{0.5}$
            & 70\% (14/20) \\

            \cmidrule(lr){3-4}
            & & $\pi_0\text{-FAST}$
            & 30\% (\hspace{4pt}6/20) \\

            \cmidrule(lr){2-4}
            & \multirow{3}{*}{Real-time}
            & $\pi_0 $ + RTC
            & 75\% (15/20) \\

            & & $\pi_{0.5}$+ RTC
            & 80\% (16/20) \\

            \cmidrule(lr){3-4}
            & &
            \cellcolor{gray!20}
            $\bm{\pi_0}\textbf{-REALFAST}$
            &
            \cellcolor{gray!20}
            80\% (16/20) \\

            \bottomrule
        \end{tabular}%
    }
    \caption{}
    \label{fig:droid_dynamic}
\end{subfigure}

\caption{\textbf{Preliminary results on various hardware configurations and dynamic tasks.} (a) Hardware-specific inference latency analysis for decoding 11 tokens. (b) Success rate of the dynamic task after additional fine-tuning with 430 demonstrations on top of pre-trained DROID policies.}

\end{figure*}

For direct comparison with prior work~\cite{black2026real}, we have followed the same hardware configuration: an RTX 4090 GPU, which may not generalize to other hardware configurations.
As a preliminary analysis, we analyze inference latency under different hardware configurations as shown in \Cref{fig:droid_hardware}.
We find that inference latency $\delta_{t, m}(N)$ increases linearly with $N$ relative to the hardware-specific pure latency $\delta_{t, m}(1)$, regardless of the hardware choice, as discussed in \Cref{sec:method_tailoring}.
However, in low-memory-bandwidth hardware scenarios that cause high inference latency, such as Jetson Orin ($\approx 200$ GB/s), changing $m$ and $N$ is not enough to achieve reasonable reactivity compared to the RTX 4090 GPU and A100 GPU.
This implies that orthogonal approaches, such as weight quantization~\citep{lin2024awq} and speculative decoding~\citep{leviathan2023fast}, are required for real-time execution in low-resource scenarios.

Meanwhile, we have mainly focused on experiments in a widely adopted static environment~\cite{khazatsky2024droid} to validate the claim in previous work~\cite{pertsch2025fast} under real-time execution.
We have extended it to zero-shot evaluation on dynamic tasks, such as picking up a moving object on a treadmill as illustrated in \Cref{fig:droid_dynamic}, but find that all policies fail on this task.
However, when fine-tuning with demonstrations that involve waiting for the moving object to approach and reacting to grasp it, we observe that $\pi_0\text{-REALFAST}$ is comparable to $\pi_0$ and $\pi_{0.5}$ with RTC and shows significant improvements over $\pi_0\text{-FAST}$.
We confirm that fast reactivity in a policy trained in static environments is insufficient, while dynamic tasks can be learned even with relatively slower reactivity through real-time execution, leaving an open question about finding the general level of reactivity required to solve dynamic tasks.


\clearpage
\section*{Acknowledgements}

This work was supported by Institute of Information \& communications Technology Planning \& Evaluation (IITP) grant funded by the Korea government(MSIT) (No.RS-2026-25516746, Development of On-Device VLA-Based Real-Time Action Intelligence Technology for Task Execution of Autonomous Agents).
This work was partly supported by the Institute of Information \& Communications Technology Planning \& Evaluation (IITP) grant funded by the Korean government (MND) (No. RS-2026-25553157, Military-Industry-Academia Cooperation Center(Seoul)), the Institute of Information \& Communications Technology Planning \& Evaluation (IITP) grant funded by the Korean government (MSIT) (No. RS-2021-II211343, Artificial Intelligence Graduate School Program (Seoul National University), No. RS-2026-25522885, Development of a World Foundation Model for Training and Development of Physical AI Systems, No. RS-2026-25512061, Development of Large Action Model-Based Autonomous Digital Twin Operation Technology for Proactive Problem Solving), the National Research Foundation of Korea (NRF) grant funded by the Korean government (MSIT) (No. RS-2024-00354218, No. RS-2026-25561904), National IT Industry Promotion Agency(NIPA) grant funded by the Korea government(MSIT) (Development of Physical Intelligence Action Models and an Integrated Framework, No.RS-2026-25621677), and the Technology Innovation Program (RS-2025-25456760, Development of a humanoid robot specialized in chemical processes based on AI foundation model) funded by the Ministry of Trade, Industry and Resources (MOTIR, Korea). 
We express special thanks to KAIT GPU project. 
The ICT at Seoul National University provides research facilities for this study.
This work was partially supported by a Google Research grant.


\bibliography{main}  

\clearpage

\appendix

\renewcommand{\thefigure}{\Alph{figure}}
\renewcommand{\thetable}{\Alph{table}}
\renewcommand{\thealgorithm}{\Alph{algorithm}}

\renewcommand{\theHfigure}{\Alph{figure}}
\renewcommand{\theHtable}{\Alph{table}}
\renewcommand{\theHalgorithm}{\Alph{algorithm}}

\setcounter{figure}{0}
\setcounter{table}{0}
\setcounter{algorithm}{0}

\section{Implementation Details}

\subsection{Constrained Decoding under FAST with Byte-level BPE Tokenization}

\begin{figure}[h]
\centering
\begin{lstlisting}[language=Python]
def prepare_C_v(v_bytes: Dict[int, List[int]]):
    all_C_v = {}
    for i, bytes in v_bytes:
        C_v = 0
        for b in bytes:
            # Exclude the continuation bytes in UTF-8 encoding
            if not (128 <= b <= 191):
                c_v = c_v + 1  
        all_C_v[i] = C_v
        
    return all_C_v

def construct_DP_table(B: int, C: int, all_C_v: Dict[int, int]):
    # We only care about the unique capacity size in C_v for T_DP
    C_v = list(set(all_C_v.values()))
    
    # DP table of achieving capacity c with exactly i decoding steps
    T_DP = np.zeros((B+1, C+1), dtype=bool)
    T_DP[0, 0] = True
    for i in range(1, B+1):
        for c in C_v:
            if c <= C:
                if c == 0:
                    T_DP[i, :] = T_DP[i, :] | T_DP[i-1, :]
                else:
                    T_DP[i, c:] = T_DP[i, c:] | T_DP[i-1, :-c] 
    # DP table of achieving capacity c with at least i decoding steps
    T_DP = np.logical_or.accumulate(T_DP, axis=0)
    
    return T_DP
\end{lstlisting}

\vspace{-6pt}
\caption{\textbf{Python-style pseudocode for $\bm{C_v}$ and $\bm{T}_\textbf{DP}$ under FAST with BBPE tokenization.} We assume that the corresponding bytes of token $v \in \mathcal{V}$ is given as \lstinline[language=Python]|v_bytes: Dict[int, List[int]]|.}

\label{fig:appendix_constrained}
\end{figure}

Throughout all experiments, we have implemented our approach using JAX based on the \texttt{openpi}\footnote{\url{https://github.com/Physical-Intelligence/openpi}} repository to adopt $\pi_0\text{-FAST}$~\citep{pertsch2025fast} as the base model for our $\pi_0\text{-REALFAST}$, including the implementation of RTC~\citep{black2026real} for $\pi_0$~\citep{black2024pi_0} and $\pi_{0.5}$~\citep{black2025pi_}, which is not supported in officially released implementation\footnote{\url{https://github.com/Physical-Intelligence/real-time-chunking-kinetix}}.
Here, we have confirmed that the official implementation of FAST adopts \textit{Byte-level} BPE (BBPE) tokenizer~\citep{wang2020neural} for token length compression rather than the BPE tokenizer~\citep{sennrich2016neural} as mentioned in the original paper.
Thus, we provide additional implementation details to be considered when implementing our approach for constrained decoding within the officially released FAST implementation, including the vectorized construction of $T_{\text{DP}}$ to help further understanding of the proposed method.

FAST considers the BPE tokenization for DCT coefficients $\vec{C}_t \in \mathbb{R}^{H \times d}$ which is the result of orthogonalized Type-II DCT is applied to the action chunk $\vec{A}_t \coloneqq [\vec{a}_{t}, \vec{a}_{t+1}, \cdots, \vec{a}_{H}]\in \mathbb{R}^{H \times d}$, where each dimension of $\vec{a}_t \in \mathbb{R}^d$ is normalized to the range $[-1, 1]$ by dataset statistics.
Specifically, each coefficient $C_{i, j}$ constituting $\vec{C}_t$ is quantized into $Q_{i, j} \coloneqq \lfloor \gamma \cdot C_{i, j} \rceil \in \mathbb{Z}$ according to the scale hyperparameter $\gamma$, and a unique occurrence of $Q_{i, j}$ in the dataset is considered as the \textit{initial alphabet} in the BPE tokenizer.
$Q_{i, j}$ are first linearized according to the action dimension, and then merged as tokens according to the BPE tokenizer starting from the initial alphabet to form the vocabulary $\mathcal{V}$.

Therefore, the token sequence $\vec{v}$ of the action chunk tokenized by FAST must have the same length with the length of the original DCT space coefficients $\vec{C}_t$ before quantization when decoded into the set of quantized coefficients $\{Q_{i, j}\}$ according to the vocabulary $\mathcal{V}$.
In other words, the capacity $C$ and $C_v$ that we have discussed correspond to $C=H \times d$ and $C_v = |\{Q_{i, j}\}_{v}|$, where $\{Q_{i, j}\}_{v}$ denotes corresponding length of quantized DCT coefficient for each token $v \in \mathcal{V}$ for FAST.
This implies that we can count the \textit{total number of characters} corresponding to each $v$ to compute $C_v$, assuming that FAST is trained with the BPE tokenizer and initial $Q_{i, j}$ is mapped to a Unicode character.

However, $C_v$ cannot be calculated based on character length when using BBPE tokenization in FAST, since BPE merge operations occur at the raw byte level rather than at the $Q_{i, j}$ level.
More precisely, $Q_{i, j}$ is mapped into a \textit{variable-length byte sequence} under UTF-8 encoding according to the index when the initial vocabulary is constructed, and the bytes constituting $Q_{i, j}$ are merged into $v$ according to their respective occurrences.
Thus, we need to isolate only the influence of the byte constituting $Q_{i, j}$ in $v$ under these byte sequences, and it can be done by measuring the length of bytes excluding \textit{continuation bytes} in UTF-8 encoding to verify the pure contribution as a character.
In other words, for coefficient values mapped outside the ASCII code range, we only count the \textit{leading bytes} for $C_v$.

\Cref{fig:appendix_constrained} demonstrates the Python-style pseudocode for computing $C_v$ in FAST with BBPE tokenization, and constructing the DP table $T_{\text{DP}}$.
Here, $C_v$ and $T_{\text{DP}}$ are precomputed before deployment and stored to enable $O(1)$ lookup for any new token $v$ during constrained decoding.
We only need to iterate over the unique values of $C_v$ during the construction of $T_{\text{DP}}$, because the update depends solely on the remaining capacity $c$.
Furthermore, we can optimize the table construction by noting that reaching a capacity $c$ in at most $i$ steps is logically equivalent to reaching it in exactly $j$ steps, for some $0 \le j \le i$.
Therefore, we first compute only the DP table for \textit{exact-step} condition $T'_{\text{DP}}$, and then fulfill the \textit{at most} condition all at once by applying the efficient vectorized cumulative OR operation:
\begin{equation}
\begin{split}
T'_{\text{DP}} [i, c] = \bigvee_{v \in \mathcal{V}} T'_{\text{DP}} [i-1, c - C_v] \rightarrow T_{\text{DP}} [i, c] = \bigvee_{j=0}^i T'_{\text{DP}} [j, c].
\end{split}
\end{equation}

\subsection{Asynchronous Inference for Real-time Execution}

For real-time execution, we assume that the controller client executes a single $\vec{a}_t$ from the action queue $\vec{A}_Q$ at every controller timing, and the policy server repeats inference asynchronously whenever an observation $\vec{o}_t$ provided from the client exists.
This is intended to ensure the policy server reacts to environmental changes as frequently as possible, reflecting the importance of inference frequency for performance as observed in the analysis of $d_m$ for the Kinetix benchmark~\citep{black2026real, matthews2025kinetix} in \Cref{sec:method_analysis}.
Therefore, we no longer consider the execution horizon $s$ specified by the client side. 
It is equivalent to always considering the case $s = d_m$ for a non-modifiable prefix length $m$, unlike RTC~\citep{black2026real}, which always considers $s_\text{min} = 25$ for $H = 50$ and $m=d_H \in \{6, 11, 16\}$ with $s = \max (s_\text{min}, d_H)$ for their real-world deployment under $\Delta t = 20$ ms using $\pi_{0.5}$.
\Cref{alg:appendix_server} and \Cref{alg:appendix_client} show the pseudocode for the policy server and the controller client implemented from the described perspective.

\vspace{-8pt}

\begin{minipage}{0.49\textwidth}
    \begin{algorithm}[H]
        \caption{Policy Server}
        \label{alg:appendix_server}
        \begin{algorithmic}[1]
            \REQUIRE  policy $\pi_\theta$, Mutex $\mathcal{M}$, observation cache $\vec{o}_t$,  action queue $\vec{A}_Q$, prefix length $m$
            
            \WHILE{True}
                \STATE \textbf{if} $\vec{o}_t$ is not None \textbf{then}
                        \STATE \quad \textbf{lock} $\mathcal{M}(\vec{o}_t, \vec{A}_Q)$
                            \STATE \qquad $\vec{o}_{t, p} \leftarrow \vec{o}_t$
                            \STATE \qquad $\vec{A}_{t,p} \leftarrow \vec{A}_Q[0:m]$
                            \STATE \qquad $\vec{o}_t \leftarrow $ None
                            \STATE \qquad $\vec{A}_Q \leftarrow \vec{A}_Q[0:m]$
                        \STATE \quad \textbf{unlock} $\mathcal{M}(\vec{o}_t, \vec{A}_Q)$
                        \STATE \quad $\vec{A}_t[m:2m] \sim \pi_\theta(\cdot \mid \vec{o}_{t, p}, \vec{A}_{t, p})$
                        \STATE \quad \textbf{lock} $\mathcal{M}(\vec{A}_Q)$
                            \STATE \qquad $\text{push}(\vec{A}_Q, \vec{A}_t[m:2m])$ 
                        \STATE \quad \textbf{unlock} $\mathcal{M}(\vec{A}_Q)$
                \STATE \textbf{end if}
            \ENDWHILE
        \end{algorithmic}
    \end{algorithm}
\end{minipage}
\hfill
\begin{minipage}{0.49\textwidth}
    \begin{algorithm}[H]
        \caption{Controller Client}
        \label{alg:appendix_client}
        \begin{algorithmic}[1]
            \REQUIRE controller $\mathcal{C}$, Mutex $\mathcal{M}$, observation cache $\vec{o}_t$, action queue $\vec{A}_Q$, prefix length $m$
            
            \WHILE{True}
                \STATE \textbf{lock} $\mathcal{M}(\vec{o}_t, \vec{A}_Q)$
                    \STATE \quad $\vec{o}_t \leftarrow \mathcal{C}$
                    \STATE \quad \textbf{if} $\vec{a}_t \ne \vec{a}_{\text{no-op}}$ \textbf{then}
                        \STATE \qquad $\text{pop}(\vec{A}_Q)$
                    \STATE \quad \textbf{end if}
                    \STATE \quad \textbf{if} $\vec{A}_Q$ is not empty \textbf{then}
                        \STATE \qquad $\vec{a}_t \leftarrow \vec{A}_Q[0]$
                    \STATE \quad \textbf{else}
                        \STATE \qquad $\vec{a}_t \leftarrow \vec{a}_{\text{no-op}}$
                    \STATE \quad \textbf{end if}
                \STATE \textbf{unlock} $\mathcal{M}(\vec{o}_t, \vec{A}_Q)$
                \STATE $\mathcal{C} \leftarrow \vec{a}_t$ when timing of $\mathcal{C}$ is reached
            \ENDWHILE
        \end{algorithmic}
    \end{algorithm}
\end{minipage}

\subsection{Reducing Action Chunk Horizon and Prefix Conditioning}

To achieve a reasonable average latency and allow separated tokenization of the action chunks, we have reduced the tokenization horizon of FAST+~\citep {pertsch2025fast} in $\pi_0\text{-FAST}$, the action tokenizer trained on 1-second-length action chunks, to $400$ ms without architectural changes.
Although this tokenization horizon differs from the pretrained horizon, we find that the reconstruction MSEs of the normalized action chunks are approximately 3e-4 and 5e-4 in the LIBERO and DROID environments, respectively, which are similar to the levels reported in the original work for 1-second-length action chunks~\citep{pertsch2025fast}.
Therefore, we simply modify the prompt template of the $\pi_0\text{-FAST}$ to support prefix conditioning as described in \Cref{fig:appendix_template} without additional consideration of reduced tokenization horizon, and only minimize the cross-entropy loss on tokens corresponding to tokenized actions and related headers.

\begin{figure}[h]
    \centering
    
    \vspace{-6pt}
    \begin{tcolorbox}[colback=gray!5!white, colframe=gray!75!black, arc=2mm, title=Prompt Template of $\pi_0\text{-FAST}$]
    Task: <instruction>, State: <quantized proprioceptive state>;\\
    \textbf{Action: < $\bm{\mathcal{T}_H(\vec{A}_t)}$ > [EOS]}
    \end{tcolorbox}

    \vspace{-4pt}
    \begin{tcolorbox}[colback=gray!5!white, colframe=gray!75!black, arc=2mm, title=Prompt Template of $\pi_0\text{-REALFAST}$]
    Task: <instruction>, State: <quantized proprioceptive state>;\\
    \textbf{Action: < $\bm{\mathcal{T}_m(\vec{A}_t[0 : m])}$ >}\\
    \textbf{Next Action:} \tcbox[on line, colback=yellow!30, colframe=yellow!30, size=fbox, arc=1mm, boxrule=0pt]{\textbf{< $\bm{\mathcal{T}_m(\vec{A}_t[m : 2m])}$ > [EOS]}}
    \end{tcolorbox}
    
    \vspace{-6pt}
    \caption{\textbf{Prompt template of \bm{$\pi_0\text{-FAST}$} and \bm{$\pi_0\text{-REALFAST}$}.} We denote the tokens for loss calculation in bold and decode only the highlighted tokens when the action chunk prefix is provided.}
    \label{fig:appendix_template}

\end{figure}

\subsection{Multi-trajectory Decoding and KV Cache Sharing}

In multi-trajectory decoding, we compute the KV caches of observation $\vec{o}_t$ and the token sequence considered as a prefix on the prompt template as in a single forward pass, without duplicating the token sequence for each trajectory~\citep{jang2025verifier}.
Here, these KV caches are not copied for each trajectory in the decoding stage but are referenced via broadcasting during parallel sampling of each trajectory, preventing additional IO latency from duplicated KV caches.
Each trajectory additionally stores the capacity accumulated so far at each decoding step to apply constrained decoding by masking the logits based on the lookup of shared DP table $T_\text{DP}$ and $C_v$, which are precomputed before deployment.
We choose the best-of-$N$ trajectory by sorting the token sequences by the accumulated token log-likelihood on the constrained support and selecting the sequence with the maximum log-likelihood, without applying additional likelihood normalization based on token length for the sake of simplicity.

\subsection{Computation Resources and Latency Measurements}

\begin{table}[h]
    \centering
    
    \vspace{-14pt}
    \caption{\textbf{Measured latency and possible selection of $\bm{m}$ in experimental setups.} The external latency $\epsilon_t$ in the LIBERO environment is assumed to restrict real-time execution during deployment.}
    
    \vspace{2pt}
    \adjustbox{max width=\columnwidth}{
        \begin{tabular}{lclccccc}
            \toprule
            \textbf{Environment} & $\Delta t$ &\textbf{Model} & $\bm{\sup \delta_{t, m}}$ & $\bm{\mathbb{E}[\epsilon_{t, m}]}$ & $\bm{m}$ \\
            \midrule
            \multirow{3}{*}{LIBERO} & \multirow{3}{*}{$100$ ms} & $\pi_0$ + RTC & $70$ ms  & \multirow{3}{*}{$\ge 100$ ms} & $\ge 1$ \\
            & & $\pi_{0.5}$ + RTC & $74$ ms & & $\ge 1$ \\
            & & $\pi_0{\text{-REALFAST}}$ & $190$ ms & & $4$ \\
            \midrule
            \multirow{3}{*}{DROID} & \multirow{3}{*}{$66.7$ ms} & $\pi_0$ + RTC & $70$ ms & \multirow{3}{*}{$9$ ms} & $\ge 1$ \\
            & & $\pi_{0.5}$ + RTC & $74$ ms & & $\ge 1$ \\
            & & $\pi_0{\text{-REALFAST}}$ & $315$ ms & & $6$ \\
            \bottomrule
        \end{tabular}
    }
    \label{tab:appendix_latency}
    
\end{table}

We use 4 NVIDIA A100 80GB GPUs and 8 NVIDIA H200 141GB GPUs for training in the LIBERO and DROID environments, respectively.
We choose a reasonable baseline for inference resources during the deployment of modern VLAs as a single NVIDIA RTX 4090 GPU, following RTC~\citep{black2026real}.
Specifically, we assume deployment on a workstation equipped with a single RTX 4090 GPU, an AMD RYZEN 7 3900X CPU, and 128GB of main memory in each environment, and measure latency $\delta_{t, m}$ with this workstation.
In \Cref{tab:appendix_latency}, we summarize the latency statistics that influence the choice of $m$ in our experimental environment.
Here, for $\pi_0$ and $\pi_{0.5}$ with RTC applied in the DROID environment, the minimum non-modifiable prefix length is $m=1$; however, we conservatively choose $m=2$ as the default to preserve consistency across the trial from unexpected latency peaks.

\subsection{Hyperparameter Configurations}

\begin{table}[h]
    \centering
    
    \vspace{-14pt}
    \caption{\textbf{Hyperparameter configurations of $\bm{\pi_0}\text{-REALFAST}$.} We report the best-performing hyperparameters of training steps and $\tau$ from the ablation study conducted in the LIBERO environment.}
    
    \vspace{2pt}
    \adjustbox{max width=\columnwidth}{
        \begin{tabular}{lcc}
            \toprule
            \textbf{Hyperparameter} & \textbf{LIBERO} & \textbf{DROID} \\
            \midrule
            Action Space & $\Delta$EEF & Cartesian velocity \\
            Optimizer & \multicolumn{2}{c}{AdamW~\citep{loshchilov2017decoupled}} \\
            $(\beta_0, \beta_1)$ & \multicolumn{2}{c}{(0.9, 0.95)}\\
            Weight Decay & \multicolumn{2}{c}{1e-8}\\
            Learning Rate Scheduler & \multicolumn{2}{c}{cosine with linear warmup}\\
            (Peak / Decay) Learning Rate & \multicolumn{2}{c}{(5e-5 / 5e-6)} \\
            Warmup Ratio & \multicolumn{2}{c}{0.1} \\
            Batch Size & \multicolumn{2}{c}{256} \\
            Training Steps & 6,000 & 300,000 \\
            Precision & \multicolumn{2}{c}{bfloat16} \\
            Gradient Clipping & \multicolumn{2}{c}{1.0} \\
            Model EMA Decay & \multicolumn{2}{c}{0.999} \\
            \midrule
            $m$ & 4 & 6 \\
            $B$ & 11 & 20 \\
            $(N, \tau)$ & \multicolumn{2}{c}{(4,  0.3)} \\
            \bottomrule
        \end{tabular}
    }
    \label{tab:appendix_hparam}
\end{table}

In \Cref{tab:appendix_hparam}, we report hyperparameter configurations of $\pi_{0}\text{-REALFAST}$ in each environment.
Here, we choose 300k training steps following $\pi_{0}\text{-FAST}$ in the DROID environment, corresponding to approximately 3 epochs after filtering idle timesteps.
We also set $\tau$ to the best-performing value from the ablation study in the LIBERO environment without further hyperparameter searches.

\section{Restriction of Real-Time Execution in the LIBERO Benchmark}
\label{app:libero_details}

The standard LIBERO benchmark can perform rollout without accounting for the pausing behavior induced by policy latency in synchronous inference.
Therefore, we perform simulated evaluations restricted to meet the requirement of real-time execution, using real-world latency statistics in \Cref{tab:appendix_latency}, forcing the policy to perform inference that accounts for the endpoint robot's execution horizon $s$.
Specifically, we set $s$ equal to the upper bound of $d_m$ based on $\delta_{t, m}$ and the given $\epsilon_t$, and the policy must satisfy the necessary condition of real-time execution under the possible choice of $m$ that does not change the $s$-length prefix of $\vec{A}_Q$, except for the initial timestep since $\vec{A}_Q$ is empty at the start.

\section{Evaluation Suites for Zero-shot DROID Deployment}
\label{app:droid_details}

In \Cref{tab:appendix_droid_tasks}, we list the 8 tasks included in the evaluation suites for policies in the zero-shot DROID deployment for unseen scenes~\citep{pertsch2025fast}.
DROID policies operate a Franka Research 3 7-DoF robot arm equipped with a Robotiq 2F-85 gripper, based on a single third-view and a wrist-view RGB camera input.
In evaluation, a policy attempts 10 trials per task with a time limit of 40 seconds, corresponding to $t=600$ controller sampling timestamps, and a trial is considered successful if the robot reaches a state aligned with the instruction, as exemplified by the goal state.
Here, we define the task throughput rate as the ratio of the remaining time from the success timestamp to the maximum time limit.

\begin{table}[ht]
    \centering
    
    \caption{\textbf{Evaluation suites for zero-shot DROID deployment.} We report observations from the third-view RGB camera for initial and goal states, while the policies also utilize a wrist RGB camera.}
    \label{tab:appendix_droid_tasks}

    \includegraphics[width=\textwidth]{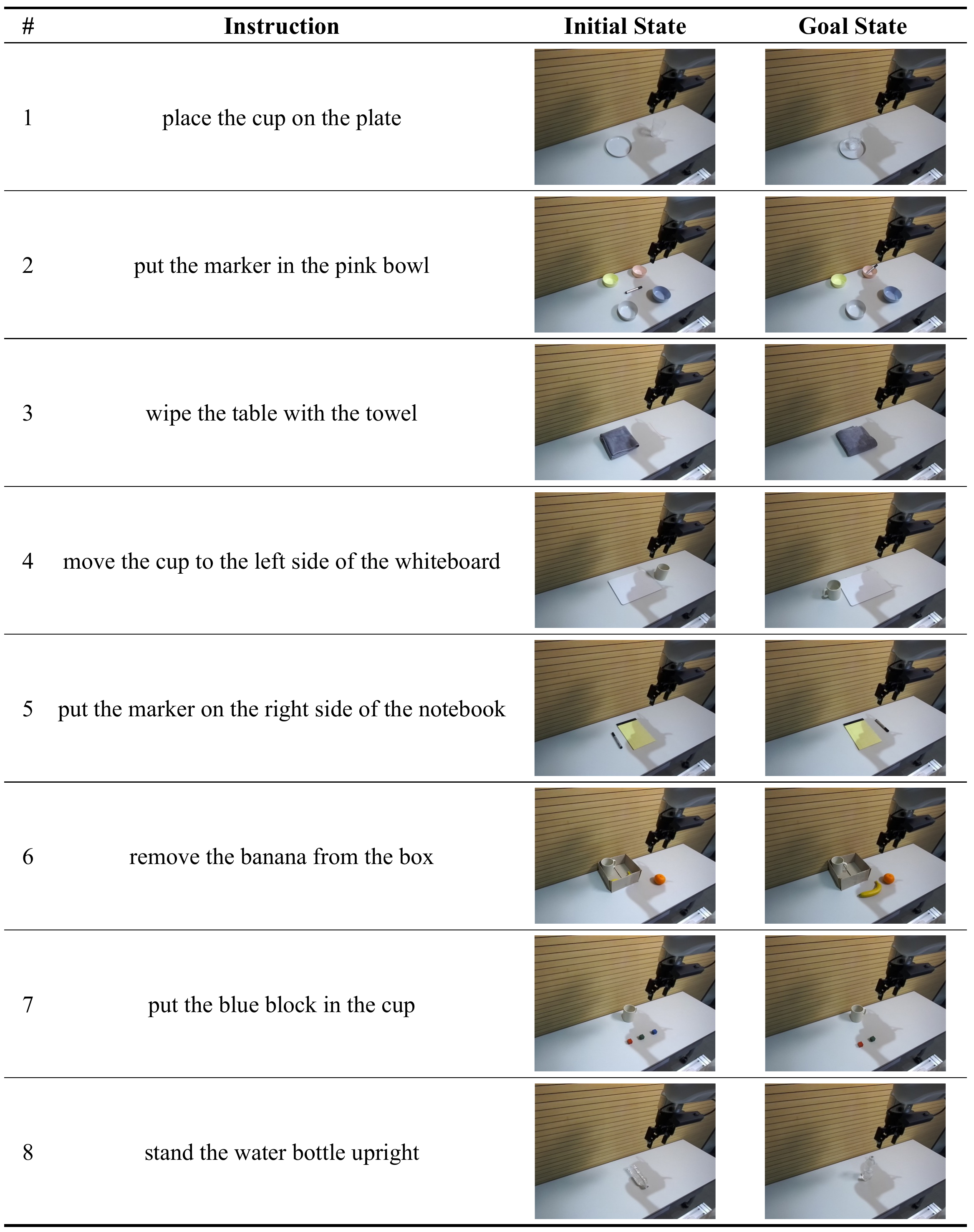}
\end{table}

\section{Task-level Evaluation Results in the Zero-shot DROID Deployment}

\begin{figure}[h]
    \centering
    
    \vspace{-4pt}
    \begin{subfigure}{0.48\columnwidth}
        \includegraphics[width=\columnwidth]{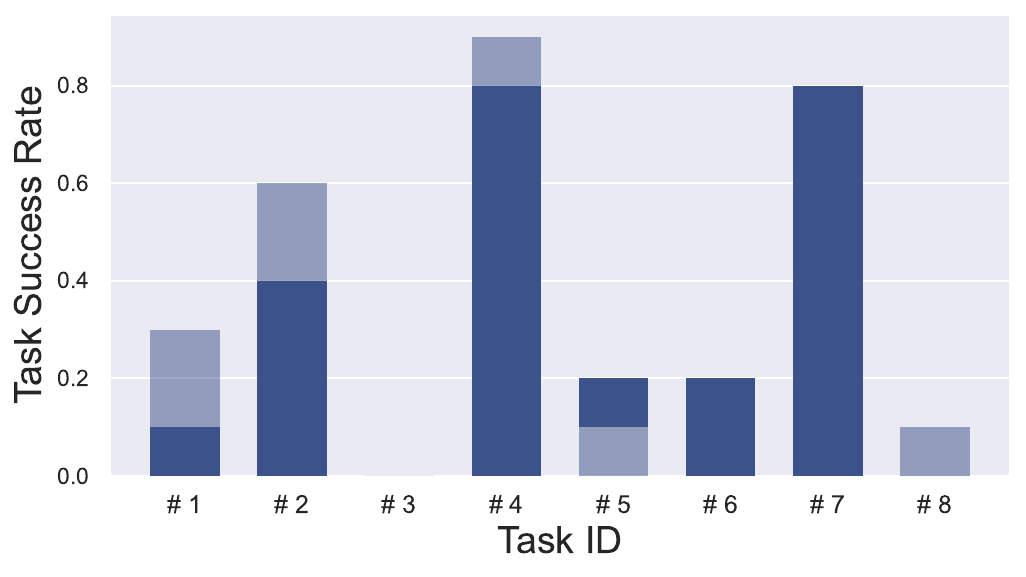}
    \end{subfigure}
    \hfill
    \begin{subfigure}{0.48\columnwidth}
        \includegraphics[width=\columnwidth]{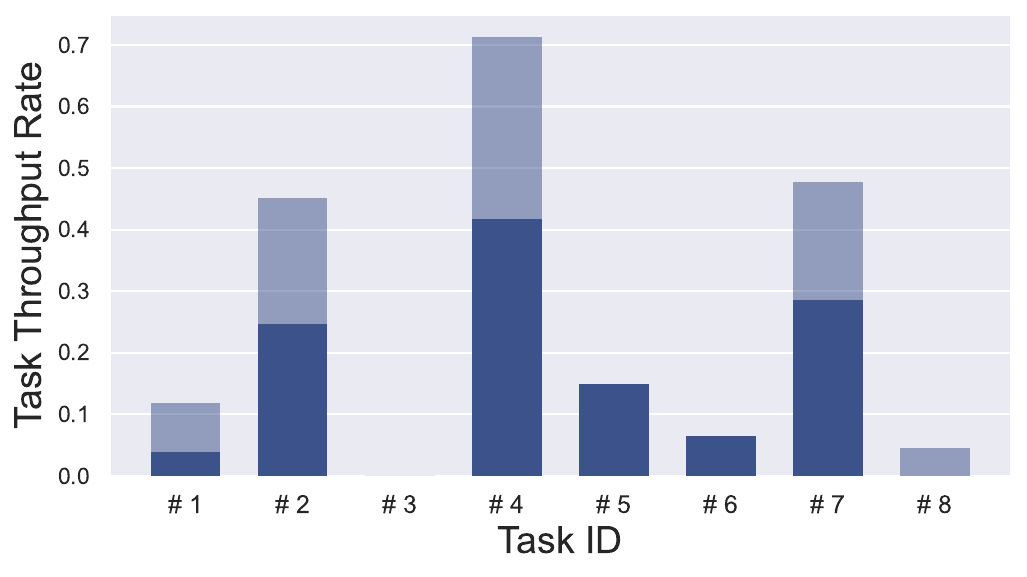}
    \end{subfigure}
    
    \begin{subfigure}{0.48\columnwidth}
        \includegraphics[width=\columnwidth]{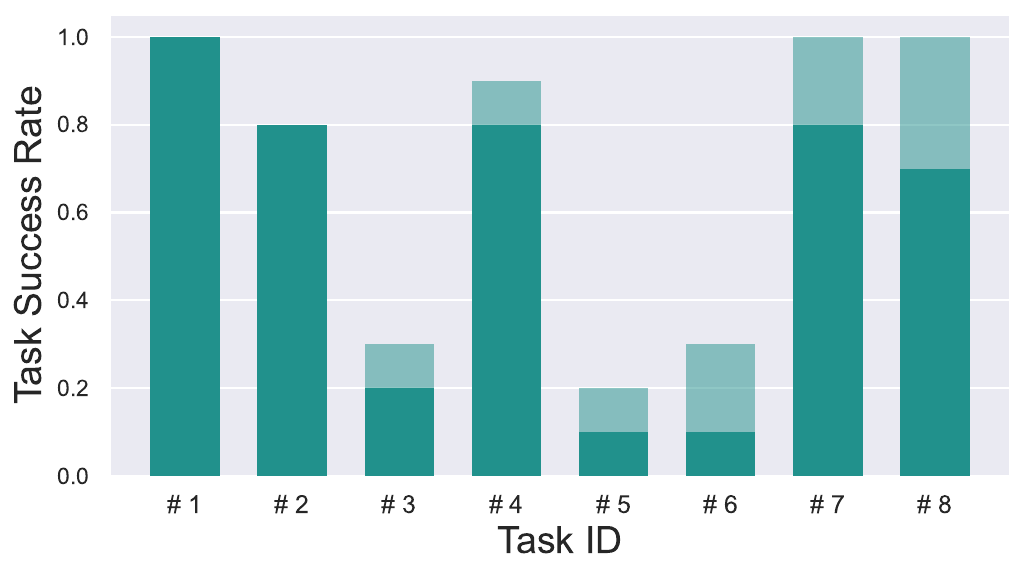}
    \end{subfigure}
    \hfill
    \begin{subfigure}{0.48\columnwidth}
        \includegraphics[width=\columnwidth]{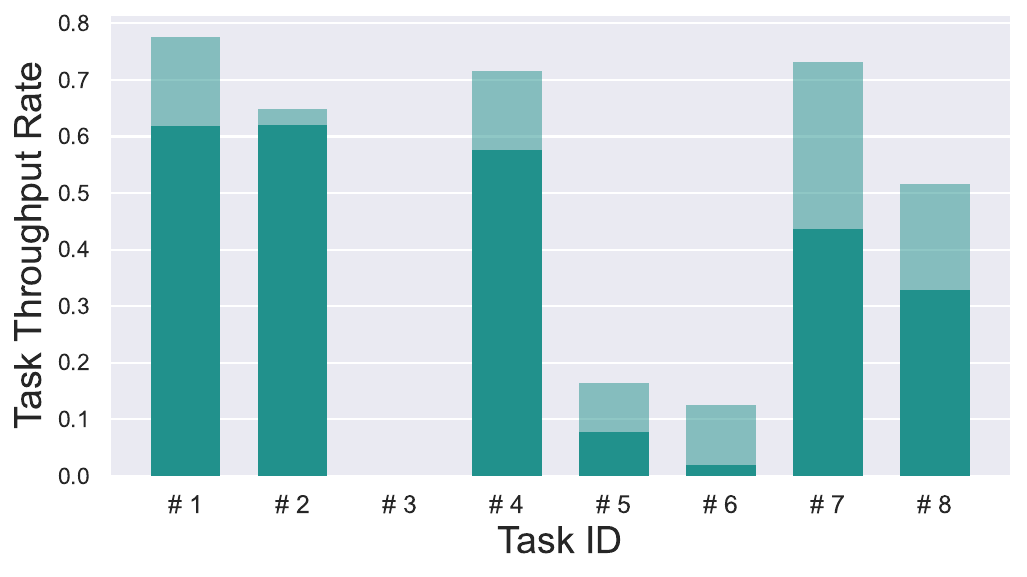}
    \end{subfigure}
    
    \begin{subfigure}{0.48\columnwidth}
        \includegraphics[width=\columnwidth]{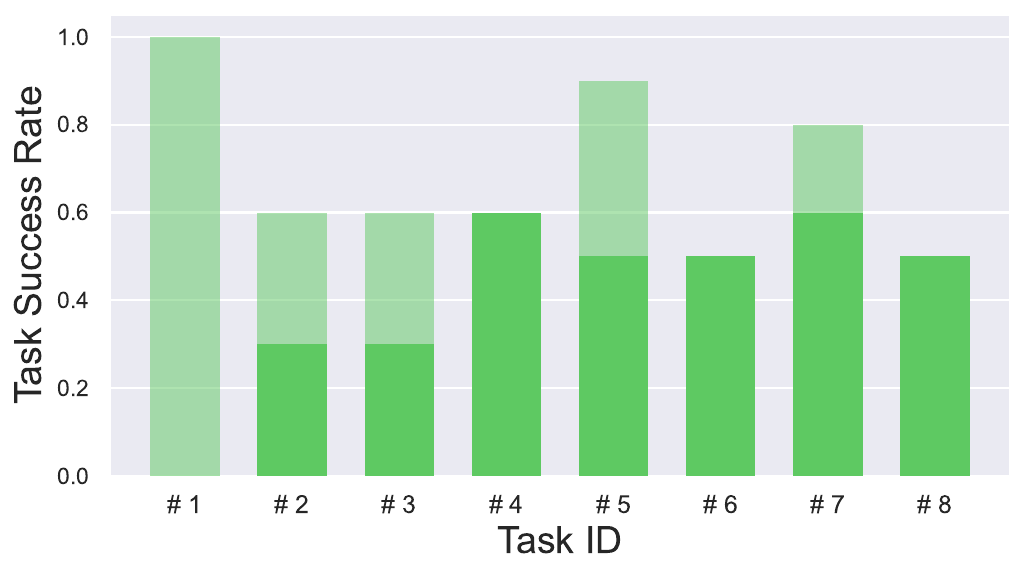}
    \end{subfigure}
    \hfill
    \begin{subfigure}{0.48\columnwidth}
        \includegraphics[width=\columnwidth]{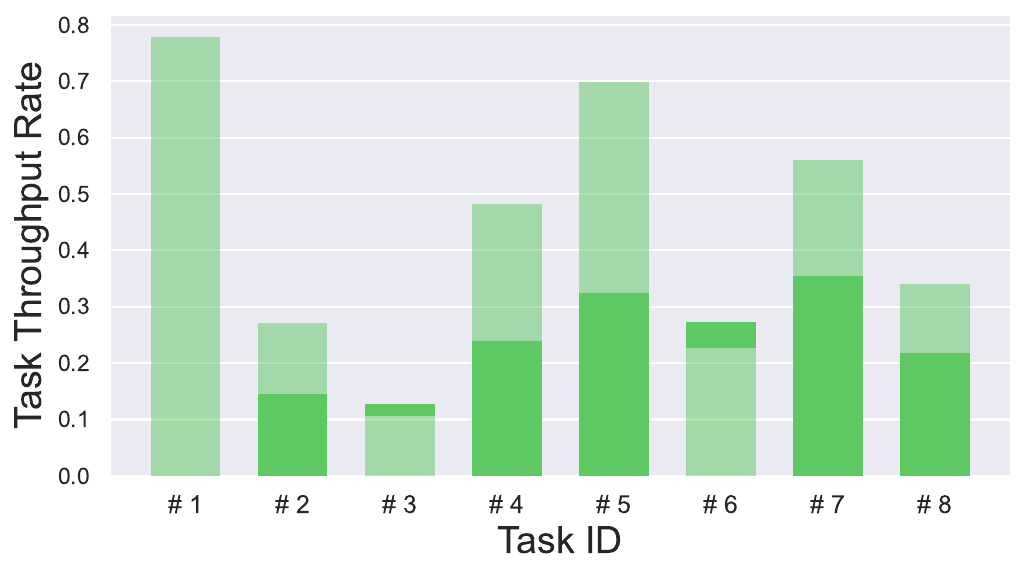}
    \end{subfigure}
    
    \vspace{-4pt}
    \caption{\textbf{Task-level evaluation results in the DROID environment.} We report synchronous inference results of $\pi_0$ (top), $\pi_{0.5}$ (middle), and $\pi_0{\text{-FAST}}$ (bottom) while the results of RTC with $m=6$ for $\pi_0$, RTC with $m=2$ for $\pi_{0.5}$, and $\pi_0{\text{-REALFAST}}$ are illustrated with lower opacities.}
    \label{fig:appendix_droid_tasks_quantitative}

\end{figure}

We perform quantitative analysis of task-level success and throughput rates in \Cref{fig:appendix_droid_tasks_quantitative}.
Here, we confirm that real-time execution generally yields improvements in task success and throughput rates for tasks that already achieve some degree of success in synchronous inference, compared with those that do not perform well in synchronous inference.
However, while there are tasks where a reasonable level of task success rate is achieved regardless of the policy (e.g., \# 4 and \# 7), we observe that there are tasks where $\pi_0\text{-FAST}$ exhibits a trend completely different from $\pi_0$ and $\pi_{0.5}$ (e.g., \# 5 and \# 6).
In particular, we identify the existence of a task where $\pi_0\text{-FAST}$ completely fails but $\pi_0\text{-REALFAST}$ succeeds (i.e. \# 1), and estimate that the major cause of this difference is the frequent action modification in real-time execution, as discussed in the qualitative comparison of \Cref{sec:experimental_results}.

\section{Further Qualitative Analysis in the DROID Environment}

\subsection{Improvement in Rollout Speed}

\begin{figure}[ht]
    \centering

    \includegraphics[width=\textwidth]{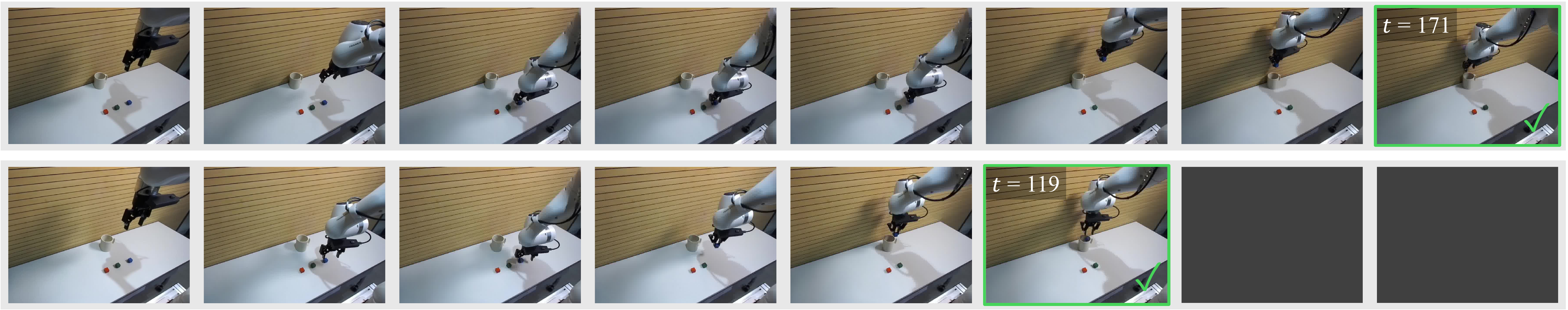}

    \vspace{-2pt}
    \caption{\textbf{Comparison of rollout speed between $\bm{\pi_0}\textbf{-FAST}$ (top) and $\bm{\pi_0}\textbf{-REALFAST}$ (bottom).} We show the trials in task \# 7 where the instruction is provided as \textit{"put the blue block in the cup"}.}
    \label{fig:appendix_droid_rollout_speed_7}
    
\end{figure}

\begin{figure}[ht]
    \centering

    \includegraphics[width=\textwidth]{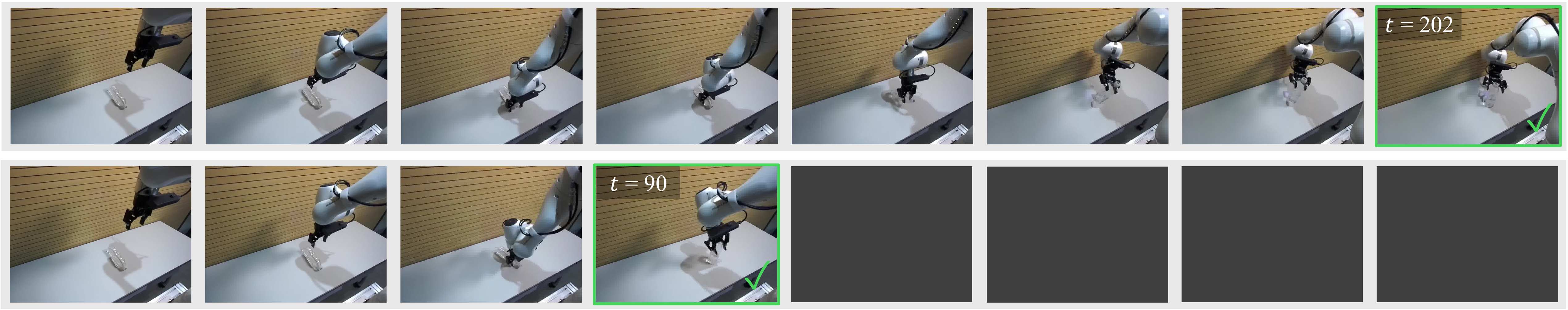}

    \vspace{-2pt}
    \caption{\textbf{Comparison of rollout speed between $\bm{\pi_0}\textbf{-FAST}$ (top) and $\bm{\pi_0}\textbf{-REALFAST}$ (bottom).} We show the trials in task \# 8 where the instruction is provided as \textit{"stand the water bottle upright"}.}
    \label{fig:appendix_droid_rollout_speed_8}
    
\end{figure}

We compare the improvement of $\pi_0\text{-REALFAST}$ over $\pi_0\text{-FAST}$ in rollout speed using the trial with the best remaining time across all trials in each task, to minimize the effect of slowed rollout speed due to recovery from mistakes.
In simple object relocation tasks where long-stride movements are affordable for task success, yet $\pi_0\text{-REALFAST}$ is generally faster in task success time compared to $\pi_0\text{-FAST}$, the best case of $\pi_0\text{-FAST}$ can reach $\pi_0\text{-REALFAST}$ so that it can marginalize the benefits of real-time execution in rollout speed as shown in \Cref{fig:appendix_droid_rollout_speed_7}.
However, we also observe that real-time execution yields significant speedups for tasks that require relatively dexterous manipulation, compared to synchronous inference, as shown in \Cref{fig:appendix_droid_rollout_speed_8}.
These examples indicate that while $\pi_0\text{-REALFAST}$ can mitigate the rollout speed bottlenecks of $\pi_0\text{-FAST}$ across overall task types, the effect of real-time execution is pronounced in task types that require fine-grained movements.


\subsection{Investigation of Failure Cases}

\begin{figure}[h]
    \centering

    \includegraphics[width=\textwidth]{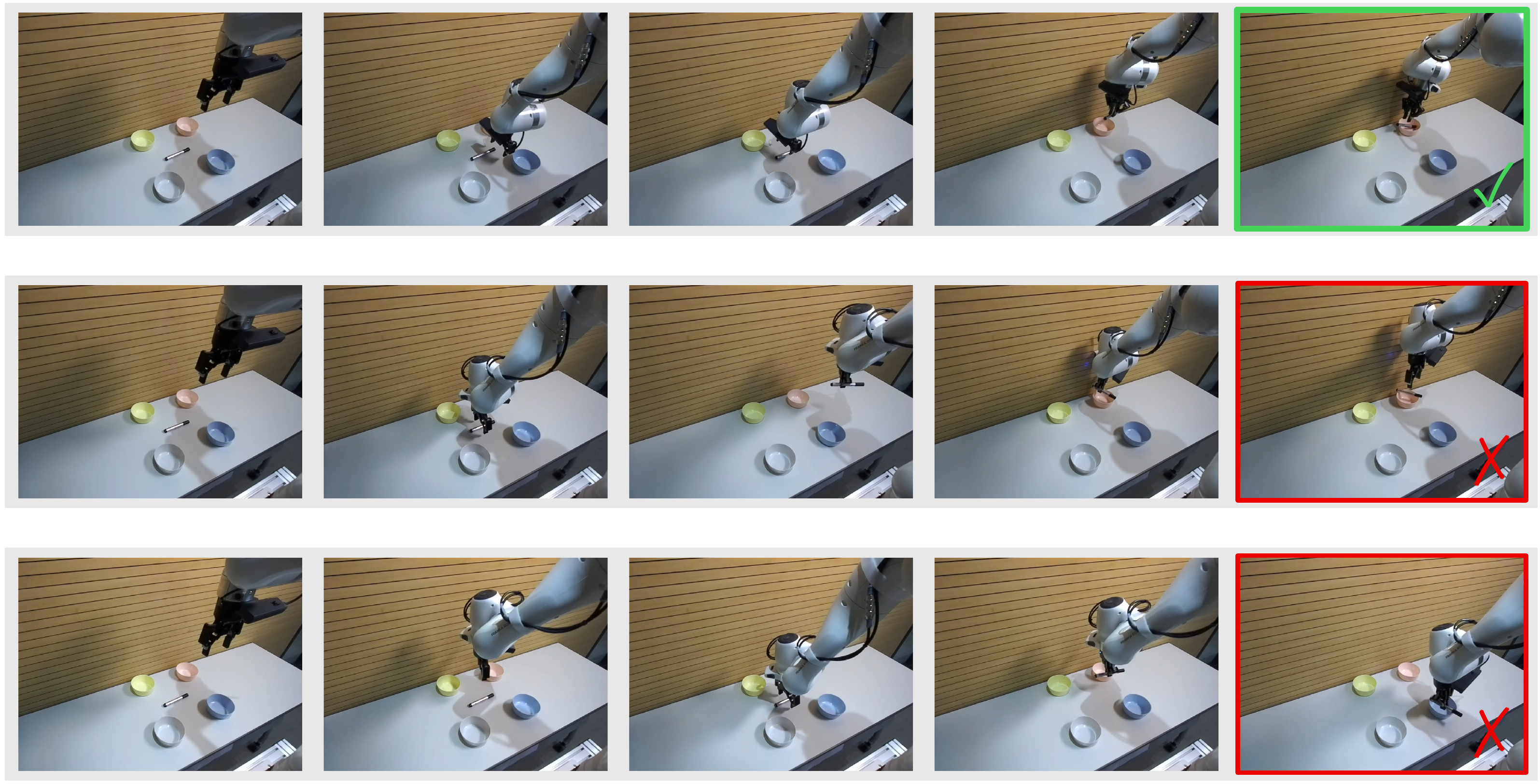}

    \vspace{-2pt}
    \caption{\textbf{Failure case analysis of $\bm{\pi_0}\textbf{-REALFAST}$.} We show a successful case (top) along with failure cases (middle, bottom) for task \# 2, where the instruction is \textit{"put the marker in the pink bowl"}.}
    \label{fig:appendix_droid_failure}
    
\end{figure}

In \Cref{fig:appendix_droid_failure}, we analyze failure cases of $\pi_0\text{-REALFAST}$, which can be semantically grounded rather than failures in basic object manipulation.
We confirm that there exist trials where $\pi_0\text{-REALFAST}$ successfully achieves the required goal state by properly following the given instruction, while there also exist cases where it achieves a goal state corresponding to a different instruction that can induce similar behavior with the original instruction.
Specifically, although the instruction is given as \textit{"put the marker in the pink bowl"}, failure cases such as placing the marker \textit{on} the pink bowl or taking the marker to the \textit{blue} bowl rather than the pink bowl are still observed.
This shows that, even though the policy already has embodied behavior for achieving task completion in the zero-shot setting, task failure still occurs due to failure in following instructions.
Therefore, it implies that further research is still required to achieve robust instruction-following of VLAs, since even autoregressive policies, which demonstrate superior generalizability in instruction-following~\cite {pertsch2025fast}, still fail at the basic level.

\end{document}